%% file: icml2020_dcg.tex
\documentclass[dvipsnames]{article}

\usepackage{microtype}
\usepackage{graphicx}
\usepackage{subfigure}
\usepackage{booktabs} 

\usepackage{url}
\usepackage{algorithm}
\usepackage[noend]{algpseudocode}
\usepackage{xcolor}
\usepackage{wrapfig}
\usepackage{enumitem}
\usepackage{multicol}

\usepackage{hyperref}

\usepackage[accepted]{icml2020/icml2020}

\usepackage{verbatim}

\icmltitlerunning{Deep Coordination Graphs}

\input{core/mathdefinitions}

\begin{document}
\setlength{\abovedisplayskip}{8pt}
\setlength{\belowdisplayskip}{8pt}

\twocolumn[
\icmltitle{Deep Coordination Graphs}



\icmlsetsymbol{equal}{*}

\begin{icmlauthorlist}
\icmlauthor{Wendelin B\"ohmer}{ox}
\icmlauthor{Vitaly Kurin}{ox}
\icmlauthor{Shimon Whiteson}{ox}
\end{icmlauthorlist}

\icmlaffiliation{ox}{Department of Computer Science, Oxford University, United Kingdom}

\icmlcorrespondingauthor{Wendelin B\"ohmer\hspace{15mm}}%
	{wendelinboehmer@gmail.com}

\icmlkeywords{Multi-agent Reinforcement Learning, Coordination Graphs}

\vskip 0.3in
]



\printAffiliationsAndNotice{}  

\input{core/abstract}
\input{core/introduction}

\input{core/background}

\input{core/method}

\input{core/related_work}
\input{core/validation}

\input{core/discussion}
\input{core/acknowledge}

\bibliography{core/references}
\bibliographystyle{icml2020/icml2020}

\appendix
\input{core/appendix}

\end{document}

%% file: core/mathdefinitions.tex
\usepackage{amsmath,amssymb,amsfonts,bm}

\def\R{\mathrm{I\kern-0.4ex R}}
\def\N{\mathrm{I\kern-0.4ex N}}
\def\E{\mathrm{I\kern-0.4ex E}}
\newcommand{\Set}[1]{\mathcal{#1}}

\newcommand{\mat}[1]{\mathbf{#1}}
\newcommand{\ve}[1]{\bm #1}

\DeclareMathOperator*{\argmax}{arg\,max}

\newcommand{\smallsum}[2]{ {\textstyle 
	\sum\limits_{\scriptscriptstyle #1}^{\scriptscriptstyle #2}} 
}

\newcommand{\smallfrac}[2]{ {\textstyle \frac{#1}{#2}} }

%% file: core/abstract.tex
\vspace{-1mm}
\begin{abstract}
	\vspace{-2mm}
	This paper introduces the \emph{deep coordination graph} (DCG)
	for collaborative multi-agent reinforcement learning.
	DCG strikes a flexible trade-off between 
	representational capacity and generalization 
	by factoring the joint value function of all agents
	according to a coordination graph
	into payoffs between pairs of agents.
	The value can be maximized by local message passing along the graph,
	which allows training of the value function end-to-end with $Q$-learning.
	Payoff functions are approximated with deep neural networks
	that employ parameter sharing and low-rank approximations
	to significantly improve sample efficiency.
	We show that DCG can solve predator-prey tasks 
	that highlight the {\em relative overgeneralization} pathology,
	as well as challenging StarCraft II micromanagement tasks.
\end{abstract}

%% file: core/introduction.tex
\section{Introduction} \label{sec:introduction}

\begin{figure*}
	\includegraphics[width=\textwidth]{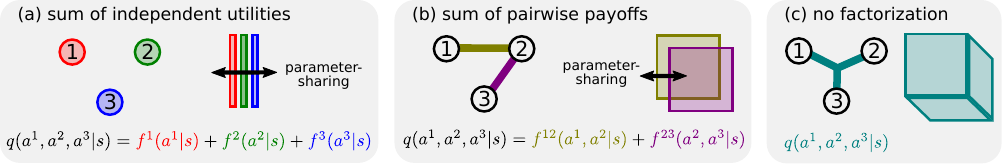}
	\vspace{-6mm}
	\caption{Examples of value factorization for 3 agents:
			(a) sum of independent utilities \citep[as in VDN,][]{Sunehag18}
			corresponds to an unconnected CG.
			QMIX uses a monotonic mixture of utilities instead of a sum \citep{Rashid18};
			(b) sum of pairwise payoffs \citep{Castellini19},  
			which correspond to pairwise edges;
			(c) no factorization \citep[as in QTRAN,][]{Son19} 
			corresponds to one hyper-edge connecting all agents.
			Factorization allows parameter sharing between factors,
			shown next to the CG,
			which can dramatically improve 
			the algorithm's sample complexity.
			}
	\label{fig:fac}
	\vspace{2mm}
\end{figure*}

One of the central challenges in cooperative 
{\em multi-agent reinforcement learning} 
\citep[MARL,][]{Jadid19}
is coping with the size of the joint action space, 
which grows exponentially in the number of agents.
For example, just eight agents, 
each with six actions,
yield a joint action space  
with more than a million actions.
Efficient MARL methods must thus generalize over large joint action spaces,
in the same way that convolutional neural networks
allow deep RL to generalize over large visual state spaces. 

MARL often addresses the issue of large joint observation and action spaces 
by assuming that the learned control policy 
is fully {\em decentralized},
that is, each agent acts independently based on its own observations only.
For example, 
Figure \ref{fig:fac}a 
shows how the joint value function can be factored
into {\em utility functions} that each depend only on the
actions of one agent \citep{Sunehag18, Rashid18}.
Consequently, the joint value function can be efficiently maximized
if each agent simply selects the action 
that maximizes its corresponding utility function.
This factorization can represent 
any deterministic (and thus at least one optimal) joint policy.  
However, that policy may not be learnable 
due to a game-theoretic pathology called 
{\em relative overgeneralization}\footnote{
	Not to be confused with the general term {\em generalization} 
	in the context of function approximation
	mentioned earlier.
} \citep{Panait06}:
during exploration other agents act randomly 
and punishment caused by uncooperative agents 
may outweigh rewards that would be achievable with coordinated actions. 
If the employed value function does not have the representational capacity 
to distinguish the values of coordinated and uncoordinated actions,
an optimal policy cannot be learned.

A higher-order value factorization can be expressed as an
undirected {\em coordination graph} \citep[CG,][]{Guestrin02},
where each vertex represents one agent
and each (hyper-) edge one \emph{payoff function}
over the joint action space of the connected agents.
Figure \ref{fig:fac}b shows a CG with pairwise edges 
and the corresponding value factorization.
Here the value depends non-trivially on
the actions of {\em all} agents,
yielding a richer representation.
Although the value can no longer be maximized by each agent individually,
the greedy action can be found using message passing along the edges
\citep[also known as \emph{belief propagation},][]{Pearl88}.
\emph{Sparse cooperative $Q$-learning} 
\citep{Kok06} applies CGs to MARL, 
but does not scale to real-world tasks,
as each payoff function
($f^{12}$ and $f^{23}$ in Figure \ref{fig:fac}b) 
is represented as a table over the state 
and joint action space of the connected agents.
\citet{Castellini19} use neural networks to 
approximate payoff functions 
in the simplified case of non-sequential one-shot games.
Moreover, neither approach shares parameters 
between the approximated payoff functions, 
so agents in each factor, represented by an edge,
must experience all corresponding action combinations.
This can require visiting a large subset of the joint action space.

While decentralization can be a requirement of the task at hand,
for example when communication between agents is impossible, 
many important applications that allow 
for centralized or distributed controllers face the same issues.
Examples are 
power, water or heat grids \citep{CorreaPosada15}, 
electronic trading \citep{Bacoyannis18}, 
computer games \citep{Vinyals19}, 
automatic factories \citep{Dotoli17}, 
drone swarms \citep{AlonsoMora17a}, 
driver simulations \citep{Behbahani19}, 
and routing of taxi fleets \citep{AlonsoMora17b}.

To address relative overgeneralization 
for centralized or distributed controllers,   
we propose the {\em deep coordination graph} (DCG),
a deep RL algorithm that scales CG for the first time to 
the large state and action spaces of modern benchmark tasks.
DCG represents the value function as a CG 
with pairwise payoffs\footnote{
	The method can be easily generalized to CGs with hyper-edges,
	i.e., payoff functions for more than 2 agents. 
} (Figure \ref{fig:fac}b) 
and individual utilities (Figure \ref{fig:fac}a).
This improves the representational capacity 
beyond state-of-the-art value factorization approaches 
like VDN \citep{Sunehag18} and QMIX \citep{Rashid18}.
To achieve scalability,
DCG employs parameter sharing between payoffs and utilities.
Parameter sharing between agents
has long been a staple of decentralized MARL.
Methods like VDN and QMIX 
condition an agent's utility on its history,
that is, its past observations and actions,
and share the parameters of all utility functions.
Experiences of one agent are thus used to train all.
This can dramatically improve the sample efficiency
compared to methods with unfactored values
\citep{Foerster16, Foerster18, Lowe17, SchroederDeWitt19, Son19}, 
which correspond to a CG  with one hyper-edge 
connecting all agents (Figure \ref{fig:fac}c).
DCG takes parameter sharing one step further
by approximating \emph{all} payoff functions
with the same neural network.
To allow unique outputs for each payoff,
the network is conditioned on a learned embedding
of the participating agents' histories.
This requires only one linear layer more than VDN
and fewer parameters than QMIX. 

DCG is trained end-to-end with deep $Q$-learning \citep[DQN,][]{Mnih15}, 
but uses message passing to coordinate greedy action selection 
between {\em all} agents in the graph.
For $k$ message passes over $n$ agents with $m$ actions each,
the time complexity of maximization
is only $\Set O(k m (n+m) |\Set E|)$,
where $|\Set E| \leq \frac{n^2 - n}{2}$ 
is the number of (pairwise) edges,
compared to $\Set O(m^n)$ for DQN without factorization.

We compare DCG's performance 
with that of other MARL $Q$-learning algorithms
in a challenging family of predator-prey tasks
that require coordinated actions
and hard StarCraft II micromanagement tasks.
In the former, DCG is the only algorithm that solves the harder tasks
and in the latter DCG outperforms state-of-the-art QMIX in some levels.
An open-source implementation of DCG and all discussed algorithms and tasks 
is available for full reproducibility\footnote{ 
\href{https://github.com/wendelinboehmer/dcg}{\tt https://github.com/wendelinboehmer/dcg}}.

%% file: core/background.tex
\section{Background} \label{sec:background}
\def\obsfun{\sigma}
\def\cg{\text{\sc cg}}
\def\vdn{\text{\sc vdn}}
\def\qmix{\text{\sc qmix}}
\def\tot{}

In this paper we assume a Dec-POMDP for $n$ agents
$\langle \Set S, \{\Set A^i\}_{i=1}^n, P, r, 
\{\Set O^i\}_{i=1}^n, \{\obsfun^i\}_{i=1}^n, 
n, \gamma \rangle$ \citep{Oliehoek16}.
$\Set S$ denotes a discrete or continuous set of environmental states and
$\Set A^i$ the discrete set of actions available to agent $i$.
At discrete time $t$,
the next state $s_{t+1} \in \Set S$ is drawn from transition kernel 
$s_{t+1} \sim P(\cdot|s_t, \ve a_t)$,
conditioned on the current state $s_t \in \Set S$ 
and joint action $\ve a_t \in \Set A := 
\Set A^1 \times \ldots \times \Set A^n$ of all agents.
A transition yields collaborative reward $r_t := r(s_t, \ve a_t)$,
and $\gamma \in [0,1)$ denotes the discount factor.
Each agent $i$ observes the state only partially by
drawing observations $o^i_t \in \Set O^i$ from
its observation kernel $o^i_t \sim \obsfun^i(\cdot|s_t)$.
The history of agent $i$'s observations $o^i_t \in \Set O^i$ 
and actions $a^i_t \in \Set A^i$ is in the following denoted as 
$\tau^i_t := (o^i_0, a^i_0, o^i_1, \ldots, o^i_{t-1}, a^i_{t-1}, o^i_t)
\in (\Set O^i \times \Set A^i)^t \times \Set O^i$.
Without loss of generality, 
this paper considers episodic tasks,
which yield episodes $(s_0, \{o_0^i\}_{i=1}^n, \ve a_0, r_0, 
\ldots, s_T, \{o_T^i\}_{i=1}^n)$ of varying (but finite) 
length $T$.

\subsection{Deep $Q$-learning} \label{sec:q_learning}
\vspace{-1.5mm}
The goal of collaborative multi-agent reinforcement learning (MARL)
is to find an optimal policy $\pi^*: \Set S \times \Set A \to [0,1]$,
that chooses joint actions $\ve a_t \in \Set A$ 
such that the expected discounted sum of future rewards is maximized.
This can be achieved by estimating the optimal $Q$-value function\footnote{
	We overload the notation $f(y|x)$ to also indicate the inputs $x$ 
	and multivariate outputs $y$ of multivariate functions $f$.
}:

$ $
\vspace{-6mm}
\begin{eqnarray} 
	q^*(\ve a|s) &:=&
		\E_{\pi^*}\!\Big[\,\smallsum{t=0}{T-1} \gamma^t r_t 
		\,\big| \!\!\begin{array}{c} 
			\scriptstyle s_0=s \\[-1mm]
			\scriptstyle \ve a_0=\ve a 
		\end{array}\!\!\Big] 
	\\ &=& \nonumber
		r(s, \ve a) + \gamma \int P(s'|s,\ve a) \, 
			\max q^*(\cdot|s') \, ds' \,.
\end{eqnarray}
The optimal policy $\pi^*(\cdot|s_t)$ chooses greedily the 
action $\ve a \in \Set A$ that maximizes the corresponding 
optimal $Q$-value $q^*(\ve a|s_t)$.
In fully observable discrete state and action spaces,
$q^*$\! can be learned in the limit from interactions 
with the environment \citep{Watkins92}.
For large or continuous state spaces,
$q^*$ can only be approximated, 
e.g., with a deep neural network $q_\theta$
\citep[DQN, ][]{Mnih15},
parameterized by $\theta$, by minimizing
the mean-squared Bellman error with gradient descent:
\begin{equation*}
		\label{eq:dqn_loss}
		\Set L_\text{\sc dqn} \,:=\, 
		\E\Big[ \smallfrac{1}{T} \!\smallsum{t=0}{T-1} \Big(
			r_t + \gamma \max q_{\bar\theta}(\cdot|s_{t+1})
			- q_\theta(\ve a_t|s_t) \Big)^{\!2}
		\Big] \,.
\end{equation*}
The expectation is estimated with episodes from 
an experience replay buffer holding 
previously observed episodes \citep{Lin92},
and $\bar\theta$ denotes the parameter of a separate 
target network, which is periodically 
replaced with a copy of $\theta$ to improve stability.
Double $Q$-learning further stabilizes training
by choosing the next action greedily
w.r.t.~the current network $q_\theta$, 
i.e., $q_{\bar\theta}(\argmax q_\theta(\cdot|s_{t+1}) | s_{t+1})$
instead of the target network $\max q_{\bar\theta}(\cdot|s_{t+1})$ 
\citep{Hasselt16}. 

In partially observable environments,
the learned policy cannot condition on the state $s_t$.
Instead, 
\citet{Hausknecht15} approximate a $Q$-function 
that conditions on the agent's history $\ve\tau_t := \{\tau^i_t\}_{i=1}^n$,
i.e., $q_\theta(\ve a|\ve\tau_t)$,
by conditioning a recurrent neural network
\citep[e.g., a GRU,][]{Chung14}
on the agents' observations $\ve o_t := (o^1_t, \ldots, o^n_t)$
and last actions $\ve a_{t-1}$,
that is, $q_\theta(\ve a|\ve h_t)$ conditions on
the recurrent network's hidden state
$h_\psi(\ve h_t|\ve h_{t-1}, \ve o_t, \ve a_{t-1})$,
$\ve h_0 = \ve 0$.

Applying DQN to multi-agent tasks quickly becomes infeasible,
due to the combinatorial growth of state and action spaces.
Moreover, DQN values
cannot be maximized without evaluating all actions.
To allow efficient maximization for MARL $Q$-learning, 
various algorithms based on value factorization
have been developed.
We derive IQL \citep{Tan93}, VDN \citep{Sunehag18}, QMIX \citep{Rashid18}
and QTRAN \citep{Son19} 
in Appendix \ref{sec:base_algos}.

\citet{Sunehag18} define the VDN value function
$q^\text{\sc vdn}(s_t, \ve a) := \sum_{i=1}^n f^i(a^i|s_t)$
and introduce parameter sharing between
the agents' utility functions $f^i(a^i|s_t) \approx f^v_\theta(a^i|\tau^i_t)$ 
to dramatically improve the sample efficiency of VDN.
The utility function $f^v_\theta$ has a fixed number 
of outputs $A := |\cup_{i=1}^n \Set A^i|$,
but agent $i$ can restrict maximization to $\Set A^i$
by setting the utilities of unavailable actions to $-\infty$. 
Specialized behavior between agents can be represented by conditioning 
$f^v_\theta$ on the agent's role,
or more generally on the agent's ID \citep{Foerster18, Rashid18}.

\subsection{Coordination graphs}
An undirected \emph{coordination graph} \citep[CG,][]{Guestrin02} 
$\Set G = \langle \Set V, \Set E \rangle$ contains 
a vertex $v_i \in \Set V$ for each agent $1 \leq i \leq n$
and a set of undirected edges $\{i,j\} \in \Set E$ 
between vertices $v_i$ and $v_j$.
The graph is usually specified before training,
but \citet{Guestrin02b} suggest 
that the graph could also depend on the state, 
that is, each state can have its own unique CG.
A CG induces a factorization\footnote{
	The normalizations $\frac{1}{|\Set V|}$ and $\frac{1}{|\Set E|}$
	are not strictly necessary but allow 
	the potential transfer of learned DCG to other topologies.
} of the $Q$-function
into {\em utility functions} $f^i$ and {\em payoff functions} $f^{ij}$
(Fig.~\ref{fig:fac}a and \ref{fig:fac}b):
\begin{equation}
	q^\cg(s_t, \ve a) := \smallfrac{1}{|\Set V|}\hspace{-1ex}
		\sum_{v^i \in \Set V} \!\! f^i(a^i|s_t)
		+ \smallfrac{1}{|\Set E|}\hspace{-2ex}
		\sum_{\{i,j\} \in \Set E} \!\!\!\!\! f^{ij}(a^i, a^j|s_t) \,. \!
\end{equation}
The special case $\Set E = \varnothing$ 
yields VDN, but each additional edge enables 
the value representation of the joint actions of a pair of agents
and can thus help to avoid relative overgeneralization.
Prior work also considers higher order coordination
where the payoff functions depend on the actions 
of larger sets of agents \citep{Guestrin02,Kok06,Castellini19},
corresponding to graphs with hyper-edges (Figure \ref{fig:fac}c).
For the sake of simplicity we restrict ourselves here to pairwise edges,
which yield at most $|\Set E| \leq \frac{1}{2} (n^2 - n)$ edges,
in comparison to up to $\frac{n!}{d! \, (n-d)!}$ hyper-edges of degree $d$.
The induced $Q$-function $q^\cg$ can be maximized locally
using \emph{max-plus}, also known as \emph{belief propagation} \citep{Pearl88}.
At time $t$ each node $i$ sends messages $\mu^{ij}_t(a^j) \in \R$ 
over all adjacent edges $\{i,j\} \in \Set E$.
In a tree topology, this message contains the maximized contributions
of the sender's sub-tree given that the receiver chooses $a^j \in \Set A^j$.
Messages can be computed locally as:
\begin{eqnarray} 
	\mu^{ij}_t(a^j) &\leftarrow& \nonumber
		\max_{a^i} \Big\{ \smallfrac{1}{|\Set V|} f^i(a^i|s_t) 
		+ \smallfrac{1}{|\Set E|} f^{ij}(a^i, a^j|s_t) 
	\\[1mm] && \hspace{6ex} \label{eq:cg_messages}
		+ \sum_{\{k,i\} \in \Set E} \mu^{ki}_t(a^i) - \mu^{ji}_t(a^i) \Big\} \,.
\end{eqnarray}
\vspace{-5mm}

This process repeats for a number of iterations,
after which each agent $i$ can locally find the action $a^i_*$ 
that maximizes the estimated joint $Q$-value $q^\cg(s_t, \ve a_*)$: 
\begin{equation} \label{eq:cg_max}
		a_*^i \;:=\; \argmax_{a^i} \Big\{ 
			\smallfrac{1}{|\Set V|} f^i(a^i|s_t) + 
			\sum_{\{k,i\} \in \Set E} \mu^{ki}_t(a^i) \Big\} \,.
\end{equation}
\vspace{-5mm}

Convergence of messages
is only guaranteed for acyclic CGs
\citep{Pearl88, Wainwright04}.
However, subtracting a normalization constant 
$c_{ij} := \sum_{a} \mu^{ij}_t(a) \,/\, |\Set A^i|$
from each message $\mu^{ij}$ before it is sent 
often leads to convergence in cyclic graphs as well
\citep{Murphy99,Crick02,Yedidia03}.
See Algorithm \ref{alg:greedy} in the appendix.

%% file: core/method.tex
\section{Method}
\label{sec:method}
\def\dcg{\text{\sc dcg}}

We now introduce the \emph{deep coordination graph} (DCG),
which learns the utility and payoff functions of 
a coordination graph $\langle \Set V, \Set E \rangle$
with deep neural networks.
In their state-free implementation,
\citet{Castellini19} learn a separate network for 
each function $f^i$ and $f^{ij}$.
However, properly approximating these $Q$-values 
requires observing the joint actions
of each agent pair in the edge set $\Set E$, 
which can be a significant subset 
of the joint action space of all agents $\Set A$.
We address this issue by focusing on an architecture 
that shares parameters across functions
and restricts them to locally available information,
i.e., to the histories of the participating agents.
DCG takes inspiration from highly scalable methods 
\citep{Yang18,Chen18}
and improves upon \citet{Kok06} 
and \citet{Castellini19} 
by incorporating the following design principles:
\begin{enumerate}[label=\roman*., itemsep=-1.5mm, topsep=-1.2mm, partopsep=0pt] 
	\item Restricting the payoffs 
		$f^{ij}(a^i, a^j|\tau^i_t, \tau^j_t)$ 
		to  local information of agents $i$ and $j$ only;
	\item Sharing parameters between all payoff 
		and utility functions through a common recurrent neural network;
	\item Low-rank approximation of joint-action payoff matrices 
		$f^{ij}(\cdot, \cdot|\tau^i_t, \tau^j_t)$ in large action spaces;
	\item Allowing transfer/generalization to different CGs 
		\citep[as suggested in][]{Guestrin02b}; and
	\item Allowing the use of privileged information like the global state during training. 
\end{enumerate}

Restricting the payoff's input (i) and sharing parameters (ii),
i.e.,
$f^i_\theta(u^i | \tau^i_t) \approx f^v_\theta(u^i|\ve h^i_t)$ 
and $f^{ij}(a^i, a^j|\tau^i_t, \tau^j_t) \approx 
f^e_\phi(a^i, a^j|\ve h^i_t, \ve h^j_t)$,
improves sample efficiency significantly.
Both utilities and payoffs share further parameters through a common RNN 
$\ve h_t^i := h_\psi(\cdot|\ve h^i_{t-1}, o^i_t, a^i_{t-1})$, 
initialized with $\ve h^i_0 := h_\psi(\cdot|\ve 0, o^i_0, \ve 0)$.
Note the difference to \citet{Castellini19},
which do not condition on the state or the agents' histories, 
and learn independent functions for each payoff.

The payoff function $f^e_\phi$ has $A^2$ outputs, 
$A:=|\!\cup_{i=1}^n \Set A^i|$,
one for each possible joint action of the agent pair.
For example, each agent in a StarCraft II map with 8 enemies 
has 13 actions \citep[SMAC,][]{Samvelyan19}, 
which yields 169 outputs of $f^e_\phi$.
As only executed action-pairs are updated during $Q$-learning,
the parameters of many outputs remain unchanged for long stretches of time,
while the underlying RNN $h_\psi$ keeps evolving.
This can slow down training and 
affect message passing.
To reduce the number of parameters and 
improve the frequency in which they are updated,
we propose a {\em low-rank approximation}\footnote{
	Similar to how singular values and vectors represent matrices.}
of the payoff (iii) with rank $K$,
similar to \citet{Chen18}:
\begin{equation}
\label{eq:low_rank}
	f^{e}_{\phi}(a^i\!,a^j|\ve h^i_t, \ve h^j_t) := 
	\smallsum{k=1}{K} \hat f^{k}_{\hat \phi}(a^i|\ve h^i_t, \ve h^j_t) \,
	\bar f^{k}_{\bar \phi}(a^j|\ve h^i_t, \ve h^j_t) . 
\end{equation}
The approximation can be computed in one forward pass 
with $2K\!A$ outputs and parameters $\phi := \{\hat\phi, \bar\phi\}$.
Note that a rank $K=\min\{|\Set A^i|, |\Set A^j|\}$
approximation does not restrict the output's expressiveness,
while lower ranks share parameters and updates
to speed up learning.

To support further research in transfer between tasks (iv),
the represented value function must generalize to new topologies
(i.e., zero-shot transfer).
This requires DCG to be invariant to reshuffling of agent indices.
We solve this by averaging payoffs computed from both agents' perspectives.\footnote{
	Permutation invariance requires the payoff matrix $\ve f^{ij}$, 
	of dimensionality $|\Set A^i| \times |\Set A^j|$,
	to be the same as $(\ve f^{ji})^\top$ with swapped inputs.
	We enforce this by taking the average of both.
	This retains the ability to learn
	asymmetric payoff matrices $\ve f^{ij} \neq (\ve f^{ij})^\top$.
} However, this paper does not evaluate (iv)
and we leave the transfer of a learned DCG 
onto different graphs/topologies to future work.
The DCG $Q$-value function is:
\begin{align} 
	&q^\dcg_{\theta\phi\psi}(\ve \tau_t, \ve a) \quad:=\quad
		\smallfrac{1}{|\Set V|} \sum_{i=1}^n \overbrace{
			f^v_\theta(a^i|\ve h^i_t) }^{f^{\text{V}}_{i, a^{\!i}}} 
	\\[-2mm] & \quad\; \nonumber
		+ \smallfrac{1}{2|\Set E|} \!\!\!\!\sum_{\{i,j\} \in \Set E}\!\!\! 
			\Big(\underbrace{ f^e_\phi(a^i, a^j|\ve h^i_t, \ve h^j_t) +
				 f^e_\phi(a^j, a^i|\ve h^j_t, \ve h^i_t)
				 	}_{f^\text{E}_{\{i,j\}\!, a^{\!i}\!\!, a^{\!j}}} \Big)\,.
\end{align}
However, some tasks allow access to privileged information 
like the global state $s_t \in \Set S$ during training (but not execution).
We therefore propose in (v) to use this information
in a {\em privileged bias function} $v_\varphi: \Set S \to \R$
with parameters $\varphi$:\hspace{-5mm}
\begin{equation} \label{eq:dcgv}
	q^\text{\sc dcg-s}_{\theta\phi\psi\varphi}(s_t, \ve \tau_t, \ve a)
	\quad:=\quad
	q^\dcg_{\theta\phi\psi}(\ve \tau_t, \ve a) + v_\varphi(s_t) \,.
\end{equation}
We call this approach DCG-S \citep[similar to VDN-S from][]{Rashid18} 
and train both variants
end-to-end with the DQN loss in Section~\ref{eq:dqn_loss}
and Double $Q$-learning \citep{Hasselt16}.
Given the tensors
(multi-dimensional arrays)
$\ve f^\text{V} \in \R^{|\Set V| \times A}$ 
and $\ve f^\text{E} \in \R^{|\Set E| \times A \times A}$,
where all unavailable actions are set to $-\infty$,
the $Q$-value can be maximized by message passing as 
defined in \eqref{eq:cg_messages} and \eqref{eq:cg_max}.
The detailed procedures of computing 
the tensors (Algorithm \ref{alg:annotate}),
the $Q$-value (Algorithm \ref{alg:qvalue})
and greedy action selection (Algorithm \ref{alg:greedy}) 
are given in the appendix.
Note that we do not 
propagate gradients through the message passing loop, 
as DQN maximizes the value computed by the target network.

The key benefit of DCG lies in its ability to prevent {\em relative overgeneralization}
during the exploration of agents:
take the example of two hunters who have cornered their prey.
The prey is dangerous and attempting to catch it alone 
can lead to serious injuries.
From the perspective of each hunter, 
the expected reward for an attack depends 
on the actions of the other agent,
who initially behaves randomly.
If the punishment for attacking alone outweighs 
the reward for catching the prey,
agents that cannot represent the value for joint actions 
(QMIX, VDN, IQL) cannot learn the optimal policy.
However, estimating a value function over the joint action
space (as in QTRAN) can be equally prohibitive,
as it requires many more samples for the same prediction quality.
DCG provides a flexible function class 
between these extremes that 
can be tailored to the task at hand.

%% file: core/related_work.tex
\section{Related Work}

\citet{Jadid19} provide a general overview of cooperative deep MARL.
Independent $Q$-learning \citep[IQL][]{Tan93} 
decentralizes the agents' policy by
modeling each agent as an independent $Q$-learner.
However, the task from the perspective of a single agent 
becomes nonstationary as other agents change their policies.  
\citet{Foerster17} show how to 
stabilize IQL when using experience replay buffers.
Another approach to decentralized agents 
is {\em centralized training and decentralized execution} \citep{Kraemer16}
with a {\em factored value function} \citep{Koller99}.
Value decomposition networks \citep[VDN,][]{Sunehag18}
perform central $Q$-learning with a value function that 
is the sum of independent utility functions for each agent 
(Figure \ref{fig:fac}a).
The greedy policy can be executed 
by maximizing each utility independently.
QMIX \citep{Rashid18} improves upon this approach 
by combining the agents' utilities with a mixing network,
which is monotonic in the utilities and depends on the global state. 
This allows different mixtures in different states
and the central value can be maximized independently 
due to monotonicity.
All of these approaches are derived in Appendix \ref{sec:base_algos}
and can use parameter sharing 
between the value/utility functions.
However, they represent the joint value with 
independent values/utilities 
and are therefore susceptible to  
relative overgeneralization.
We demonstrate this by comparing DCG with all the above algorithms.

Another straightforward way to decentralize in MARL
is to define the joint policy as a product of independent agent policies.
This lends itself to the actor-critic framework,
where the critic is discarded during execution 
and can therefore condition on the global state
and all agents' actions during training.
Examples are MADDPG \citep{Lowe17} for continuous actions and 
COMA \citep{Foerster18} for discrete actions.
\citet{Wei18} specifically investigate 
relative overgeneralization in continuous multi-agent tasks
and show improvement over MADDPG 
by introducing policy entropy regularization.
MACKRL \citep{SchroederDeWitt19} follows the approach in \citet{Foerster18},
but uses {\em common knowledge}
to coordinate agents during centralized training.
\citet{Son19} define QTRAN,
which also has a centralized critic
but uses a greedy actor w.r.t.\ a VDN factorized function.
The corresponding utility functions are distilled 
from the critic under constraints
that ensure proper decentralization.
\citet{Boehmer19} present another approach to decentralize 
a centralized value function,
which is locally maximized by coordinate ascent
and decentralized by training IQL agents from the same replay buffer.
Centralized joint $Q$-value functions
do not allow parameter sharing to the same extent as value factorization,
and we compare DCG to QTRAN to demonstrate 
the advantage in sample efficiency.  
Nonetheless, 
DCG value factorization can in principle 
be applied to any of the above centralized critics
to equally improve sample efficiency
at the same cost of representational capacity.

Other work deals with huge numbers of agents,
which requires additional assumptions to reduce 
the sample complexity. 
For example, 
\citet{Yang18} introduce {\em mean-field multi-agent learning} (MF-MARL), 
which factors a tabular value function for hundreds of agents
into pairwise payoff functions
between neighbors in a uniform grid of agents.
These payoffs share parameters similar to DCG.
\citet{Chen18} introduce a value factorization (FQL) 
for a similar setup based on a 
low-rank approximation of the joint value.
This approach is restricted by uniformity assumptions
between agents, but otherwise uses parameter sharing similar to DCG.
The value function cannot be maximized globally 
and must be locally maximized with coordinate ascent.
These techniques are designed for much larger sets of agents 
and the specific assumptions and design choices 
are not well suited to the tasks considered in this paper.
To demonstrate this, 
we compare against a low-rank joint value decomposition (called {\tt LRQ}), 
similar to \citet{Chen18}.

Coordination graphs (CG)
have been extensively studied in multi-agent robotics
with given payoffs \citep[e.g.][]{Rogers11,Yedidsion18}.
\citet{VanDerPol16} learn a pairwise payoff function
for traffic light control of connected intersections with DQN,
which is used for all edges in a CG of intersections.
Sparse cooperative $Q$-learning
\citep[SCQL,][]{Kok06} uses CG in discrete state and action spaces
by representing all utility and payoff functions as tables.
However, the tabular approach restricts practical application of SCQL 
to tasks with few agents and small state and action spaces.
\citet{Castellini19} use neural networks to approximate payoff functions, 
but only in non-sequential games, 
and require a unique function for each edge in the CG.
DCG addresses for the first time the question 
how CG can efficiently solve tasks with large state and action spaces,
by introducing parameter sharing between all payoffs (as in VDN/QMIX), 
conditioning on local information (as in MF-MARL)
and using low-rank approximation of the payoffs' outputs (as in FQL).

\input{core/fig_coordination}

Graph Neural Networks \citep[GNN,][]{Battaglia18}
are architectures to approximate functions on annotated graphs
by {\em learning} some message passing over the graph's edges. 
GNN can thus be used to estimate a joint Q-value in MARL, 
which conditions on graphs annotated with the observations of all agents.
In contrast to the fixed message passing of CG, 
however, 
GNN would have to {\em learn} the joint maximization 
required for Q-learning,
which would require additional losses 
and might not be feasible in practice. 
Current MARL literature uses GNN therefore either 
as independent (IQL) value functions \citep{Jiang18,Luo19} 
or as a joint critic in actor-critic frameworks 
\citep{Tacchetti18, Liu19, Malysheva19}.

%% file: core/fig_coordination.tex
\begin{figure*}[t!]
	\centering
	\includegraphics[width=0.9\textwidth]{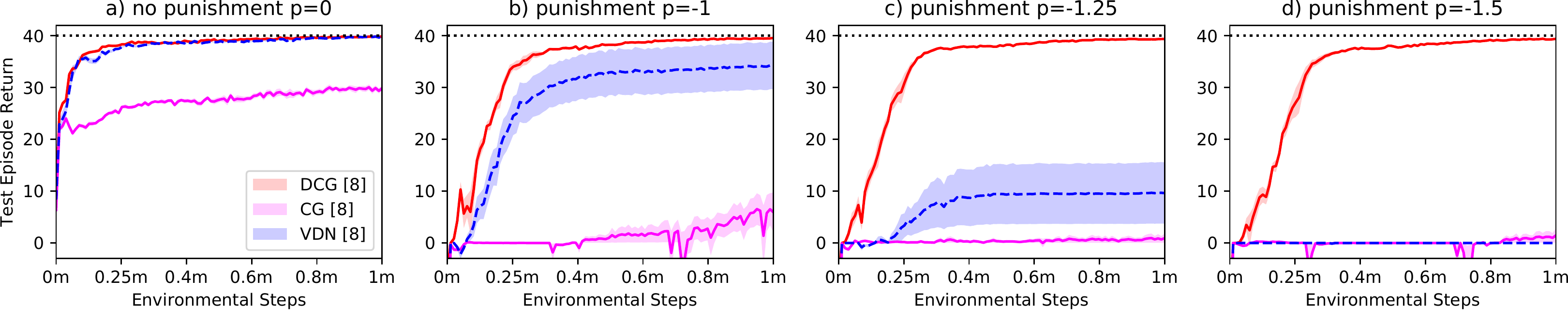}
	\vspace{-2mm}
	\caption{\label{fig:coordination}
			Influence of punishment $p$ 
			for attempts to catch prey alone
			on greedy test episode return 
			(mean and shaded standard error, [number of seeds])
			in the {\em relative overgeneralization} task 
			where 8 agents hunt 8 prey 
			(dotted line denotes best possible return).
			Fully connected DCG (\texttt{DCG})
			are able to represent the value of joint actions, 
			which leads to a better performance for larger $p$,
			where DCG without edges (\texttt{VDN}) 
			has to fail eventually ($p < -1$).
			CG without parameter sharing (\texttt{CG}),
			learn very slowly due to sample inefficiency.}
	\vspace{2mm}
\end{figure*}

%% file: core/validation.tex
\section{Empirical Results}

\input{core/tab_topologies}
\input{core/fig_coord_compare}

In this section we compare the performance of DCG
with various topologies (see Table \ref{tab:topologies})  
to the state-of-the-art algorithms {\tt QTRAN} \citep{Son19},
{\tt QMIX} \citep{Rashid18}, {\tt VDN} \citep{Sunehag18} 
and {\tt IQL} \citep{Tan93}.
We also compare a CG baseline without parameter sharing 
between payoffs and utilities
\citep[{\tt CG}, an extension of][]{Castellini19},
and a low-rank approximation of the joint value of all agents
\citep[{\tt LRQ}, similar to][]{Chen18}
that is maximized by coordinate ascent. 
Both baselines condition on a shared RNN that summarizes all agents' histories.
Lastly, we investigate how well the DCG algorithm performs
without any parameter sharing ({\tt DCG\,(nps)}\!).
All algorithms are implemented in the \textsc{pymarl} framework \citep{Samvelyan19}; 
a detailed description can be found in the appendix.

We evaluate these methods in two complex grid-world tasks
and challenging Starcraft II micromanagement tasks from the 
StarCraft Multi-Agent Challenge \citep[SMAC,][]{Samvelyan19}.
The first grid-world task formulates 
 {\em relative overgeneralization} 
as a family of predator-prey tasks
and the second investigates how {\em artificial decentralization} 
can hurt tasks that demand non-local coordination between agents.
In the latter case, 
decentralized value functions (QMIX, VDN, IQL) 
cannot learn coordinated action selection
between agents that cannot see each other directly
and thus converge to a suboptimal policy.
StarCraft II presents a challenging real-world problem
with privileged information during training,
and we compare DCG and DCG-S
on 6 levels with varying complexity.

\subsection{Relative Overgeneralization}
\label{sec:rel_overgen}
To model relative overgeneralization,
we consider a partially observable grid-world predator-prey task:
8 agents have to hunt $8$ prey in a $10 \times 10$ grid.
Each agent can either move in one of the 4 compass directions,
remain still, or try to catch any adjacent prey.
Impossible actions, i.e., 
moves into an occupied target position
or catching when there is no adjacent prey,
are treated as unavailable.
The prey moves by randomly selecting one available movement
or remains motionless if all surrounding positions are occupied.
If two adjacent agents execute the {\em catch} action,
a prey is caught and both the prey and the catching agents
are removed from the grid.
An agent's observation is a $5 \times 5$ sub-grid centered around it,
with one channel showing agents and another indicating prey.
Removed agents and prey are no longer visible 
and removed agents receive a special observation of all zeros.
An episode ends if all agents have been removed or after $200$ time steps.
Capturing a prey is rewarded with $r=10$, 
but unsuccessful attempts by single agents 
are punished by a negative reward $p$.
The task is similar to one proposed by \citet{Son19}, 
but significantly more complex, 
both in terms of the optimal policy
and in the number of agents.

To demonstrate the effect of relative overgeneralization,
Figure \ref{fig:coordination} shows the average return
of greedy test episodes for varying punishment $p$
as mean and standard error over 8 independent runs.
In tasks without punishment 
($p=0$ in Figure \ref{fig:coordination}a), 
fully connected DCG ({\tt DCG}, solid)
performs as well as DCG without edges ({\tt VDN}, dashed).
However, for stronger punishment VDN becomes more and more unreliable,
which is visible in the large standard errors 
in Figures \ref{fig:coordination}b and \ref{fig:coordination}c, 
until it fails completely for $p\leq-1.5$ in Figure \ref{fig:coordination}d.
This is due to relative overgeneralization,
as VDN cannot represent the values 
of joint actions during exploration. 
Note that a coordination graph ({\tt CG}),
where utilities and payoffs condition on all agents' observations,
{\em can} represent the value
but struggles to learn the task without parameter sharing.
DCG, on the other hand, 
converges reliably to the optimal solution (dotted line).

Figure \ref{fig:coord_compare}a shows how well DCG performs
in comparison to the baseline algorithms in Appendix \ref{sec:base_algos} 
for a strong punishment of $p=-2$.
Note that {\tt QMIX}, {\tt IQL} and {\tt VDN} 
completely fail to learn the task (return $0$)
due to their restrictive value factorization.
While {\tt CG} could in principle learn the same policy as {\tt DCG},
the lack of parameter sharing hurts performance 
as in Figure \ref{fig:coordination}.
{\tt QTRAN} estimates the values with a centralized function,
which conditions on all agents' actions,
and can therefore learn the task.
However, {\tt QTRAN} requires more samples before
a useful policy can be learned than {\tt DCG},
due to the size of the joint action space.
This is in line with the findings of \citet{Son19},
which required significantly more samples to learn 
a task with four agents than with two 
and also show the characteristic dip in performance with more agents.
{\tt LRQ} can also represent the joint value
but learns extremely slow and with large deviations 
due to imperfect maximization by coordinate ascend.
In comparison with {\tt QTRAN}, {\tt CG} and {\tt LRQ},
fully connected DCG ({\tt DCG}) learns near-optimal policies
quickly and reliably.

We also investigate the performance of various 
DCG topologies defined in Table \ref{tab:topologies}. 
Figure \ref{fig:coord_compare}b shows that in particular 
the {\em reliability} of the achieved test episode return 
depends strongly on the graph topology.
While all seeds of fully connected DCG succeed ({\tt DCG}),
DCG with {\tt CYCLE}, {\tt LINE} and {\tt STAR} topologies 
have varying means with large standard errors.
The high deviations are caused by some runs finding near-optimal policies, 
while others fail completely (return 0).
One possible explanation is that for the failed seeds
the rewarded experiences, observed in the initial exploration,
are only amongst agents that do not share a payoff function.
Due to relative overgeneralization,
the learned greedy policy no longer explores {\em catch} actions
and existing payoff functions cannot
experience the reward for coordinated actions anymore.
It is therefore not surprising 
that fully connected graphs perform best,
as they represent the largest function class 
and require the fewest assumptions. 
The topology also has little influence on the runtime of DCG,
due to efficient batching on the GPU.

The tested fully connected DCG 
only considers pairwise edges.
Hyper-edges between more than two agents
(Figure \ref{fig:fac}c)
would yield even richer value representations,
but would also require more samples 
to sufficiently approximate the payoff functions.
This effect can be seen in the slower learning
{\tt QTRAN} and {\tt LRQ} results in Figure \ref{fig:coord_compare}a.

\input{core/fig_ghosts}

\subsection{Artificial Decentralization}
\label{sec:art_decentral}

The choice of decentralized value functions
is in some cases purely artificial:
it is motivated by the huge joint action spaces 
and not because the task actually requires decentralized execution.
While this often works surprisingly well,
we want to investigate how existing algorithms deal with tasks
that cannot be fully decentralized.
One obvious case in which decentralization must fail 
is when the optimal policy cannot be represented by utility functions alone.
For example, decentralized policies behave sub-optimally
in tasks where the optimal policy would condition 
on multiple agents' observations in order to achieve the best return. 
Payoff functions in DCG, on the other hand, condition on 
pairs of agents and can thus represent a richer class of policies.
Note that dependencies on more agents can be modeled
as hyper-edges in the DCG (Figure \ref{fig:fac}c),
but this hurts the sample efficiency as discussed above.

We evaluate the advantage of a richer policy class
with a variation of the above predator-prey task.
To disentangle the effects of relative overgeneralization,
in this task prey can be caught by only one agent (without punishment).
Unbeknownst to the agent, however,
a fair coin toss decides at each time step
whether catching the prey is rewarding ($r\!=\!1$) or punishing ($r\!=\!-1$).
The current coin flip can be observe through an additional feature,
which is placed in a random corner at the beginning of each episode.
Due to the short visibility range of the agents,
the feature is only visible in one of the $9$ positions 
closest to its corner. 

Figure \ref{fig:ghosts}a shows the performance
of {\tt QTRAN}, {\tt QMIX}, {\tt IQL} and {\tt VDN}, 
all of which have decentralized policies,
in comparison to fully connected {\tt DCG}
and {\tt CG}.
The baseline algorithms have to learn a policy 
that first identifies the location of the indicating feature
and then herds prey into that corner, 
where the agent is finally able to catch it without risk.
By contrast, 
DCG and CG can learn a policy where one agent finds the indicator, 
allowing all agents that share an edge
to condition their payoffs on that agent's current observation.
As a result, this policy can catch prey much more reliably, 
as seen in the high performance of {\tt DCG} 
compared to all baseline algorithms.
Interestingly, as {\tt CG} conditions on all agents' histories simultaneously,
the baseline shows an advantage in the beginning
but then learns more slowly and reaches a significantly lower performance.
The joint value of {\tt QTRAN} conditions on all observations,
but the algorithm's constraints enforce the greedy policy to be
consistent with a VDN factorized value function, 
which appears to prevent a good performance. 
{\tt LRQ}'s factorization architecture 
appears too unstable to learn anything here.
We also investigate the influence of the DCG topologies 
in Table \ref{tab:topologies},
shown in Figure \ref{fig:ghosts}b. 
While other topologies do not reach the same
performance as fully connected DCG,
they still learn a policy that significantly outperforms 
all baseline algorithms, around the same performance
as fully connected {\tt CG}.

\subsection{Low-Rank Approximation}

While the above experiments already show a significant advantage of
DCG with independent payoff outputs for each action pair,
we observe performance issues on StarCraft II maps 
with this architecture.
The most likely cause is the difference in the number of actions per agent:
predator-prey agents choose between $|\Set A^i|=6$ actions,
whereas SMAC agents on comparable maps with 8 enemies 
have $|\Set A^i| = 13$ actions.
While payoff matrices with 36 outputs 
in predator-prey appear reasonable to learn,
169 outputs in StarCraft II would require significantly more 
samples to estimate the payoff of each joint-action properly.

Figures \ref{fig:coord_compare}c and \ref{fig:ghosts}c show the influence 
of {\em low-rank payoff approximation} 
(Equation \ref{eq:low_rank}, $K \in \{1, \ldots, 4\}$) 
on the predator-prey tasks from previous subsections.
Figure \ref{fig:coord_compare}c shows that 
any low-rank approximation (\texttt{DCG (rank $K$)}) significantly improves 
the sample efficiency over the default architecture
with independent payoffs for each action pair (\texttt{DCG (full)}).
The improvement in Figure \ref{fig:ghosts}c
is less impressive, but shows even
rank $K=1$ approximations (\texttt{DCG (rank 1)}) 
 perform slightly better than \texttt{DCG (full)}.

\subsection{Scaling Up to StarCraft II}
The default architecture of DCG
with independent payoffs for each action pair
performs poorly in StarCraft II.
We therefore test a $K\!=\!1$ low-rank payoff approximation DCG 
with (\texttt{DCG-S}) and without (\texttt{DCG}) 
privileged information bias function $v_\varphi$, 
defined in \eqref{eq:dcgv},
on six StarCraft II maps
\citep[from SMAC,][]{Samvelyan19}.
We report all learning curves in Figure \ref{fig:all_starcraft} 
of the appendix 
and show as an example 
the \emph{super hard} map \texttt{MMM2} in Figure \ref{fig:mmm2}.

\input{core/fig_starcraft_wrap}
DCG is expected to yield an advantage 
on maps that struggle with
relative overgeneralization, 
which should prevent \texttt{VDN} from learning.
We observe on almost all maps 
that \texttt{DCG} and \texttt{VDN} perform similar. 
Also, adding privileged information improves performance for
\texttt{DCG-S} and the corresponding \texttt{VDN-S} \citep{Rashid18}
in many cases. 

We conclude from the results presented in the appendix  
that, in all likelihood, the SMAC benchmark does 
not suffer from relative overgeneralization.
However, the fact that \texttt{DCG-S} matches \texttt{QMIX}, 
the state-of-the-art on SMAC,
demonstrates that the algorithm scales to complex domains like StarCraft II.
Furthermore, \texttt{DCG} and \texttt{DCG-S}
perform comparable to their \texttt{VDN} counterparts.
This demonstrates that the added payoffs and message passing, 
which allowed to overcome relative overgeneralization
in Section \ref{sec:rel_overgen},
do not affect the algorithm's sample efficiency.
This is a clear advantage over prior CG methods
\citep[\texttt{CG} in Figure \ref{fig:coordination},][]{Castellini19}.

%% file: core/tab_topologies.tex
\begin{table}[b]
	\centering
	\def\arraystretch{1.3}
	\begin{tabular}{|c|l|} 
	\hline
		{\tt DCG} & $\Set E := \big\{ \{i, j\} \,\big|\, 1 \leq i < n, i < j \leq n \big\}$ \\
		{\tt CYCLE} & $\Set E := \big\{\{i, (i \!\!\mod\! n) + 1\} \,\big|\, 1 \leq i \leq n\big\}$ \\
		{\tt LINE}  & $\Set E := \big\{\{i,i+1\} \,\big|\, 1 \leq i < n\big\}$ \\ 
		{\tt STAR}  & $\Set E := \big\{\{1, i\} \,\big|\, 2 \leq i \leq n \big\}$ \\
		{\tt VDN}  & $\Set E := \varnothing$
	\\ \hline
	\end{tabular}
	\vspace{-2mm}
	\caption{Tested graph topologies for DCG.}
	\label{tab:topologies}
\end{table}

%% file: core/fig_coord_compare.tex
\begin{figure*}[t]
	\vspace{1mm}
	\includegraphics[width=\textwidth]{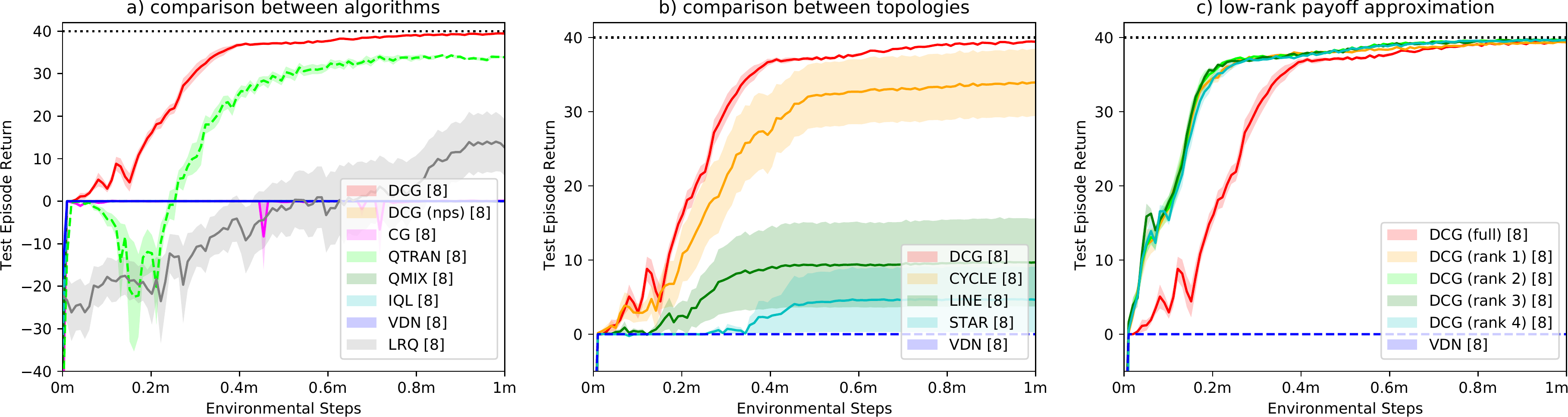}
	\vspace{-5mm}
	\caption{\label{fig:coord_compare}
			Greedy test episode return 
			for the coordination task of Figure \ref{fig:coordination}
			with punishment $p=-2$.
			Comparison (a) to baseline algorithms;
			(b) between DCG topologies;
			(c) of different low-rank payoff approximations.
			Note that {\tt QMIX}, {\tt IQL} and {\tt VDN} (dashed) 
			do not solve the task (return 0) 
			due to {\em relative overgeneralization}.
			{\tt CG}, {\tt QTRAN} and {\tt LQR} ($K=64$) 
			{\em could} represent the joint value,
			but are sample inefficient due to the large joint action spaces.
			Note that without parameter sharing, 
			{\tt DCG\,(nps)} suffers the same fate. 
			The reliability of DCG depends on the CG-topology:
			all seeds with fully connected {\tt DCG} solved the task,
			but the high standard error for {\tt CYCLE}, 
			{\tt LINE} and {\tt STAR} topologies 
			is caused by some seeds succeeding
			while others fail completely. 
			Low-rank approximation ({\tt DCG\,(rank\,$K$)}) 
			dramatically improves sample efficiency 
			without any significant impact on performance.}
	\vspace{0mm}
\end{figure*}

%% file: core/fig_ghosts.tex
\begin{figure*}[t!]
	\includegraphics[width=\textwidth]{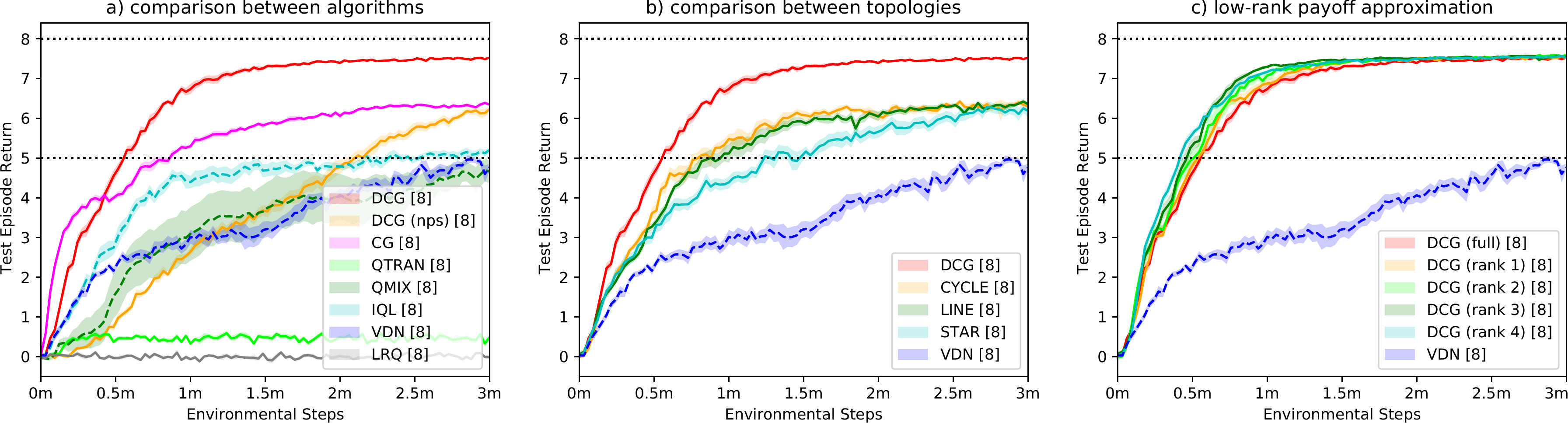}
	\vspace{-5mm}
	\caption{\label{fig:ghosts}
			Greedy test episode return 
			(mean and shaded standard error, [number of seeds])
			in a {\em non-decentralizable task} where 8 agents hunt 8 prey:
			(a) comparison to baseline algorithms;
			(b) comparison between DCG topologies;
			(c) comparison of low-rank payoff approximations.
			The prey turns randomly into punishing ghosts,
			which are indistinguishable from normal prey.
			The prey status is only visible at an indicator that is
			placed randomly at each episode in one of the grid's corners.
			{\tt QTRAN}, {\tt QMIX}, {\tt IQL} and {\tt VDN}
			learn decentralized policies, 
			which are at best suboptimal in this task (around lower dotted line).
			Fully connected {\tt DCG} and {\tt CG} can learn a near-optimal policy
			(upper dotted line denotes best possible return),
			but without parameter sharing {\tt DCG\,(nps)} and {\tt CG}
			yield sub-optimal performance in comparison to {\tt DCG}. 
			In this task low-rank approximations 
			only marginally increase sample efficiency.}
	\vspace{0mm}
\end{figure*}

%% file: core/fig_starcraft_wrap.tex
\setlength{\columnsep}{3mm}
\def\wrapscale{0.28\textwidth}
\begin{wrapfigure}{r}{\wrapscale}	
	\includegraphics[width=\wrapscale]{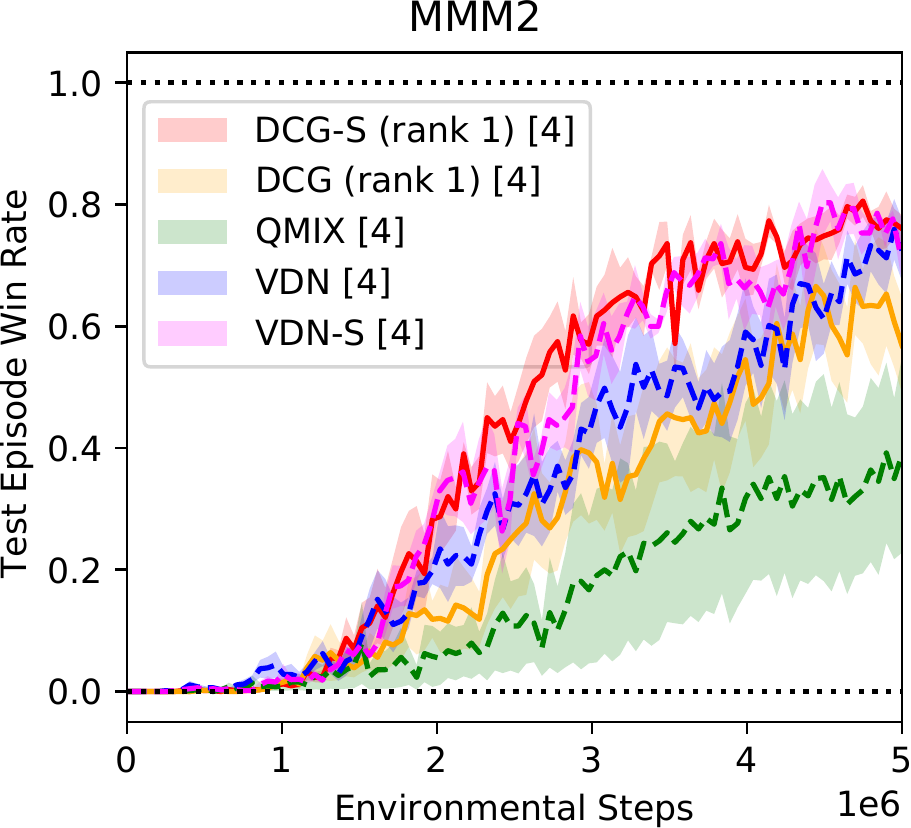}
	\vspace{-4mm}
	\caption{\label{fig:mmm2}
			Win rate of test episodes on the SMAC map \texttt{MMM2}.
			Both \texttt{DCG} and \texttt{DCG-S} use rank 1 approximation.}
	\vspace{-1mm}
\end{wrapfigure}

%% file: core/discussion.tex
\section{Conclusions \& Future Work}

This paper introduces the {\em deep coordination graph} (DCG),
an architecture for value factorization that 
is specified by a {\em coordination graph} (CG)
and can be maximized by message passing.
We evaluate deep $Q$-learning with DCG 
and show that the architecture enables learning of
tasks where {\em relative overgeneralization}
causes all decentralized baselines to fail,
whereas centralized critics are much less sample efficient than DCG.
We also demonstrate that artificial decentralization
can lead to suboptimal behavior in all compared methods except DCG.
Our method significantly improves over existing CG methods
and allows for the first time to 
use CG in tasks with large state and action spaces.
Fully connected DCG performed best in all experiments
and should be preferred in the absence of prior knowledge about the task.
The computational complexity of this  topology scales quadratically,
which is a vast improvement over the exponential scaling of joint value estimates.
Additionally, we introduce a 
low-rank payoff approximation for large action spaces
and a privileged bias function (DCG-S).
Evaluated on StarCraft II micromanagement tasks, 
DCG-S performs competitive with the state-of-the-art QMIX.
DCG can also be defined with hyper-edges 
that connect more than two agents.
Similar to our LRQ baseline, low-rank approximation 
can be used to approximate the payoff of high-order hyper-edges,
and coordinate ascend can maximize them locally. 
Furhtermore, due to its permutation invariance,
DCG has the potential to transfer/generalize to different graphs/topologies.
This would in principle allow the training of DCG on 
dynamically generated graphs 
\citep[e.g.~using an attention mechanism,][]{Liu19}.
By including hyper-edges with varying degrees,
one could allow the agents to flexibly decide in each state 
with whom they want to coordinate.
We plan to investigate this in future work. 

%% file: core/acknowledge.tex
\subsubsection*{Acknowledgments}
The authors would like to thank Tabish Rashid for his QTRAN implementation.
This project has received funding from the European Research Council (ERC) under
the European Union’s Horizon 2020 research and innovation programme (grant agreement number
637713), the National Institutes of Health (grant agreement number R01GM114311), EPSRC/MURI
grant EP/N019474/1, the JP Morgan Chase Faculty Research Award and a
generous equipment grant from NVIDIA.
Vitaly Kurin has also been supported by the Samsung R\&D Institute UK studentship.

%% file: core/appendix.tex
\appendix
\section{Appendix}

\subsection{Baseline algorithms} \label{sec:base_algos}
All discussed algoorithms are implemented 
in the PyMARL framework \citep{Samvelyan19}
and can be found at
\url{https://github.com/wendelinboehmer/dcg}.

\paragraph{IQL} {\em Independent Q-learning} \citep{Tan93}
is a straightforward approach of value decentralization 
that allows efficient maximization
by modeling each agent as an independent DQN $q^i_\theta(a^i|\tau^i_t)$.
The value functions can be trained without 
any knowledge of other agents,
which are assumed to be part of the environment. 
This violates the stationarity assumption of $P$ 
and can become therefore instable \citep[see e.g.][]{Foerster17}.
IQL is nonetheless widely used in practice, 
as parameter sharing between agents 
can make it very sample efficient.

Note that parameter sharing requires access to 
privileged information during training, 
called {\em centralized training and decentralized execution} \citep{Foerster16}.
This is particularly useful for actor-critic methods like 
MADDPG \citep{Lowe17}, Multi-agent soft Q-learning \citep{Wei18},
COMA \citep{Foerster18} and MACKRL \citep{SchroederDeWitt19},
where the centralized critic can 
condition on the underlying state $s_t$
and the joint action $\ve a_t \in \Set A$.

\paragraph{VDN}
Another way to exploit centralized training 
is {\em value function factorization}.
For example, value decomposition networks \citep[VDN,][]{Sunehag18}
perform centralized deep $Q$-learning on a joint $Q$-value function
that factors as the sum of independent {\em utility functions} $f^i$, 
for each agent $i$:\!\!
\begin{equation}
	q^\vdn_\theta(\ve\tau_t,\ve a)
	\quad:=\quad \smallsum{i=1}{n} \,  f^i_\theta(a^i|\tau^i_t) \,.
\end{equation} 
This value function $q^\vdn$ can be maximized 
by maximizing each agent's utility $f^i_\theta$ independently.

\paragraph{QMIX} \citep{Rashid18} improves upon this concept
by factoring the value function as 
\begin{equation*}
	q^\qmix_{\theta\phi}(s_t, \ve\tau_t, \ve a)
	\; := \; 
	\varphi_\phi\big(s_t, f^1_\theta(a^1|\tau^1_t), 
		\ldots, f^n_\theta(a^n|\tau^n_t)\big) \,.
\end{equation*}
Here $\varphi_\phi$ is a {\em monotonic mixing hypernetwork}
with non-negative weights that retains monotonicity in the inputs $f^i_\theta$.
Maximizing each utility $f^i_\theta$ therefore 
also maximizes the joint value $q^\qmix$,
as in VDN.
The mixing parameters are generated by a neural network,
parameterized by $\phi$, 
that condition on the state $s_t$, 
allowing different mixing of utilities in different states.
QMIX improves performance over VDN, 
in particular in  
StarCraft II micromanagement tasks \citep[SMAC, ][]{Samvelyan19}.

\paragraph{QTRAN}
Recently \citet{Son19} introduced QTRAN,
which learns the centralized critic
of a greedy policy w.r.t.~a VDN factorized function,
which in turn is distilled from the critic 
by regression under constraints.
The algorithm defines three value functions 
$q^\vdn$, $q\tot$ and $v\tot$,
where $q\tot(\ve\tau_t, \ve a)$ 
is the centralized Q-value function, 
as in Section \ref{sec:q_learning}, and 
\begin{equation}
	v\tot(\ve\tau_t) 
	\quad:=\quad 
	\max q\tot(\ve\tau_t, \cdot) 
	- \max q^\vdn(\ve\tau_t, \cdot) \,.
\end{equation}
They prove that the greedy policies w.r.t.~$q\tot$ and $q^\vdn$
are identical under the constraints:
\begin{equation} \label{eq:qtran_constraint}
	q^\vdn(\ve\tau_t, \ve a) - q(\ve\tau_t, \ve a) 
	+ v(\ve\tau_t) \quad\geq\quad 0 \,, 
\end{equation}
$\forall \ve a \in \Set A \,,
\forall \ve\tau_t \in 
\{ (\Set O^i \times \Set A^i)^t \times \Set O^i \}_{i=1}^n$, 
with strict equality if and only if 
\,$\ve a = \argmax q^\vdn(\ve\tau_t, \cdot)$\,.
QTRAN minimizes the parameters $\phi$ 
of the centralized asymmetric value $q_\phi^i(a^i| \ve\tau_t, \ve a^{-i})$, 
$\ve a^{-i} := (a^1, \ldots, a^{i-1}, a^{i+1}, \ldots, a^n)$, 
for each agent \citep[which is similar to][]{Foerster18}
with the combined loss $\Set L_\text{\sc td}$:
\begin{equation*}
\label{eq:qtran_td_loss}
	\Set L_\text{\sc td} :=
		\E\Big[\, \smallfrac{1}{nT} \smallsum{t=0}{T-1} \, \smallsum{i=1}{n} \Big(
			 r_t + \gamma \, \bar y_{t+1}^i - 
			 q^i_\phi(a^i_t|\ve\tau_t, \ve a_t^{-i}) \Big)^{\!2} \, 
		\Big] \,,
\end{equation*}
where $\bar y_{t}^i := q^i_{\bar\phi}(\bar a^i_{t}|\ve\tau_{t}, \bar{\ve a}^{-i}_{t})$
denotes the centralized asymmetric value 
and $\bar{\ve a}_{t+1} := \argmax q^\vdn_\theta(\ve\tau_{t+1}, \cdot), \forall t,$
denotes what a greedy decentralized agent would have chosen.
The decentralized value $q_\theta^\vdn$ and the greedy difference $v_\psi$,
with parameters $\theta$ and $\psi$ respectively,
are  distilled by regression of the each $q_\phi^i$
in the constraints.
First the equality constraint:
\begin{equation*} \label{eq:L_opt}
	\Set L_\text{\sc opt} 
	:= \E\Big[\, \smallfrac{1}{n(T\!+\!1)} 
		\smallsum{t=0}{T} \, \smallsum{i=1}{n} \Big(
			q_\theta^\vdn(\ve\tau_t, \bar{\ve a}_t)
			- \bot \bar y_t^i
			+ v_\psi(\ve\tau_t)
		\Big)^{\!2} \, \Big] \,,
\end{equation*}
where the `detach' operator $\bot$ stops the gradient flow through $q^i_\phi$.
The inequality constraints are more complicated.
In principle one would have to compute a loss for every
action which has a negative left hand side in \eqref{eq:qtran_constraint}.
\citet{Son19} suggest to only constraint executed actions $\ve a_t$:
\begin{align}
	\Set L_\text{\sc nopt} 
		:= &\E\Big[\, \smallfrac{1}{n(T\!+\!1)} 
		\smallsum{t=0}{T} \, \smallsum{i=1}{n} \Big(
			\min \Big\{0, 
	\\[-1mm] \nonumber &
			q_\theta^\vdn(\ve\tau_t, \ve a_t) 
			- \bot q^i_{\phi}\tot(a_t^i|\ve\tau_t, \ve a_t^{-i}) 
			+ v_\psi(\ve\tau_t) \Big\}
		\Big)^{\!2} \, \Big] \,.
\end{align}
We use this loss,
called {\tt QTRAN-base},
which performed better in our experiments 
than {\tt QTRAN-alt} \citep[see][]{Son19}.
The losses are combined to $\Set L_\text{\sc qtran} 
:= \Set L_\text{\sc td}
+ \lambda_\text{\sc opt} \, \Set L_\text{\sc opt} 
+ \lambda_\text{\sc nopt} \, \Set L_\text{\sc nopt}$ \,,
with $\lambda_\text{\sc opt}, \lambda_\text{\sc nopt} > 0$.

\paragraph{CG} To compare the effect of parameter sharing 
and restriction to local information in DCG,
we evaluate a variation of \citet{Castellini19}
that can solve sequential tasks.
In this baseline all agents share a RNN encoder of their belief 
over the current global state $\ve h_t := 
h_\psi(\cdot| \ve h_{t-1}, \ve o_t, \ve a_{t-1})$
with $\ve h_0 := h_\psi(\cdot| \ve 0, \ve o_0, \ve 0)$,
as introduced in Section \ref{sec:q_learning}.
However, the parameters of the utility or payoff functions
are not shared, that is, $\theta := \{\theta_i\}_{i=1}^n$
and $\phi := \{\phi_{ij} | \{i,j\} \in \Set E\}$.
Each set of parameters $\theta_i$ and $\phi_{ij}$
represents one linear layer from $\ve h_t$
to $\Set A^i$ and $\Set A^i \times \Set  A^j$ outputs, respectively.
Other wise the baseline uses the same code as DCG,
that is, Algorithms \ref{alg:annotate}, \ref{alg:qvalue} and \ref{alg:greedy}.

\begin{figure}[t!]
	\includegraphics[width=\linewidth]{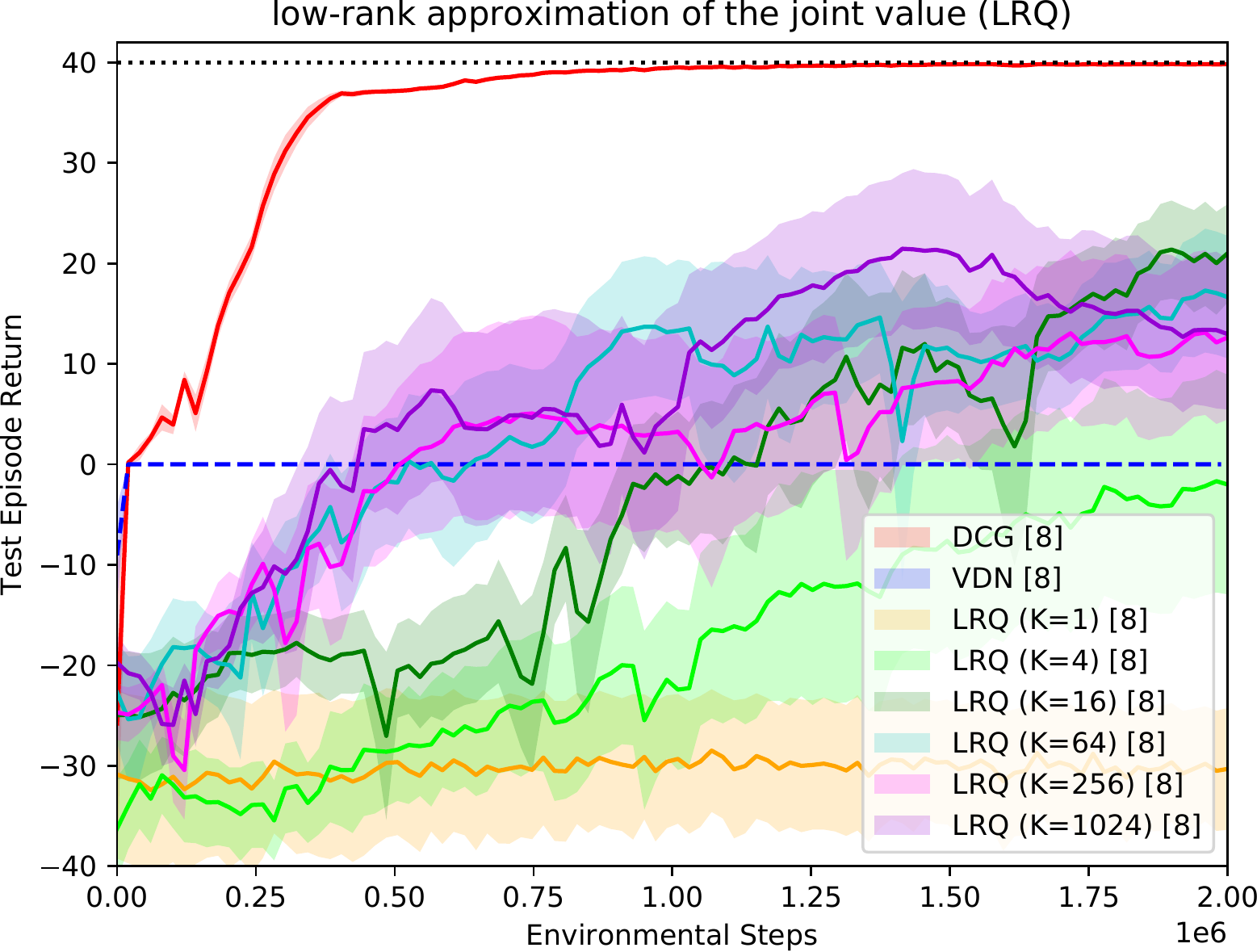}
	\vspace{-5mm}
	\caption{\label{fig:lrq}
			Low-rank approximation of the joint value function (LRQ)
			in the relative overgeneralization 
			task of Section \ref{sec:rel_overgen}
			for varying numbers of factors $K \in \{1,4,16,64,256,1024\}$.
			}
\end{figure}

\paragraph{LRQ}
As DCG uses low-rank approximation of the payoff outputs,
it is a fair question how a low-rank approximation  
of the full joint value function (LRQ) would perform
(akin to one hyper-edge shown in Figure \ref{fig:fac}c).
This approach is similar to FQL \citep{Chen18},
but we drop here the homogeneity assumptions between agents.
Instead, we define the joint value function as a sum of $K$ factors, 
which each are the product of $n$ factor functions $\bar f^{ik}$, 
one for each agent:
\begin{equation}
	q^\text{\sc lrq1}(\ve \tau_t, \ve a)
	\;:=\;
	\sum_{k=1}^K \prod_{i=1}^n 
	\bar f^{ik}_\theta(a^i \,| \ve \tau_t) \,.
\end{equation}
The joint histories of all agents $\ve \tau_t$
are encoded with a common RNN with 512 hidden neurons.
In difference to DCG, 
LQR cannot be maximized by message passing.
Instead we perform coordinate ascend
by choosing a random joint action $\bar{\ve a}_0$
and iterating, $\forall i \in \{1,\ldots,n\}$,
\begin{equation}
	\bar a^i_{l+1} \;:=\;
	\argmax_{a' \in \Set A^i}	\sum_{k=1}^K \bar f^{ik}(a'|\ve\tau_t) 
	\prod_{j\neq i} \bar f^{jk}(\bar a^j_l|\ve \tau_t) \,.
\end{equation}
The iteration finishes if the value 
$q^\text{\sc lrq1}(\ve\tau_t, \bar{\ve a}_l)$
no longer increases of after a maximum of $l=8$ iterations.

Experiments on the predator-prey tasks 
with $K \in \{1, 4, 16, 64, 256, 1024\}$ revealed 
that, due to the large input space of $\ve \tau_t$,
the above approximation did not learn anything.
To allow a better comparison, 
we use the same input restrictions 
and parameter sharing tricks as DCG,
that is, we restrict the input of each factor function to 
the history of the corresponding agent 
and share all agents' parameters:
\begin{equation}
	\ve h^i_0 \;:=\; \ve 0 \,, \qquad 
	\ve h_t^i  \;:=\; h_\psi(\cdot|\ve h^i_{t-1}, o^i_t, a^i_{t-1}) 
\end{equation}
\vspace{-4mm}
\begin{equation}
	q^\text{\sc lrq2}(\ve \tau_t, \ve a) \;:=\;
	\sum_{k=1}^K \prod_{i=1}^n 
	\bar f^k_\theta(a^i \,| \ve h^i_t) \,.
\end{equation}
Figures \ref{fig:coord_compare} and \ref{fig:ghosts}
show that this architecture learns the task with $K=64$, albeit slowly.
Figure \ref{fig:lrq} demonstrates the effect of
the number of factors $K$ on the solution
of the relative overgeneralization task of Section \ref{sec:rel_overgen}.
Given enough factors, LRQ learns the task,
albeit slowly and with a lot of variance between seeds,
probably due to imperfect maximization by coordinate ascend.

\subsection{DCG Algorithms}
All algorithms defined in this paper are given 
in pseudo-code on Page \pageref{alg:annotate}: 
Algorithm \ref{alg:annotate} computes the utility and payoff tensors,
which are used by Algorithm \ref{alg:qvalue} to compute the joint $Q$-value 
and by Algorithm \ref{alg:greedy} to return the joint actions
that greedily maximize the joint $Q$-value.
An open-source python implementation  
within the PyMARL framework \citep{Samvelyan19} 
can be found online at \qquad
\url{https://github.com/wendelinboehmer/dcg}.

\subsection{Hyper-parameters} \label{sec:hyper_parameters}
All algorithms are implemented in the \textsc{pymarl} framework
\citep{Samvelyan19}.
We aimed to keep the hyper-parameters 
close to those given in the framework and
consistent for all algorithms.

All tasks used discount factor $\gamma=0.99$
and $\epsilon$-greedy exploration,
which was linearly decayed from $\epsilon=1$
to $\epsilon=0.05$ within the first $50,000$ time steps.
Every 2000 time steps we evaluated 20 
greedy test trajectories with $\epsilon=0$.
Results are plotted by first applying 
histogram-smoothing (100 bins) to each seed,
and then computing the mean and standard error between seeds.

All methods are based on agents' histories, 
which were individually summarized with $h_\psi$
by conditioning a linear layer of 64 neurons
on the current observation and previous action,
followed by a ReLU activation 
and a GRU \citep{Chung14} of the same dimensionality.
Both layers' parameters are shared amongst agents,
which can be identified by a one-hot encoded ID in the input.
For the \texttt{CG} baseline,
the linear layer and the GRU  had $64n=512$ neurons.
This allows a fair comparison with DCG 
and also had the best final performance 
amongst tested dimensionalities $\{64, 256,512,1024\}$
in the task of Figure \ref{fig:ghosts}.
Independent value functions $q^i_\theta$ (for \texttt{IQL}), 
utility functions $f^v_\theta$ (for \texttt{VDN/QMIX/QTRAN/DCG}) 
and payoff functions $f^e_\phi$ (for \texttt{DCG})
are linear layers from the GRU output to the corresponding number of actions.
The hyper-network $\varphi_\phi$ of \texttt{QMIX} produces a mixing network 
with two layers connected with an ELU activation function, 
where the weights of each mixing-layer are 
generated by a linear hyper-layer with 32 neurons
conditioned on the global state,
that is, the full grid-world.
For \texttt{QTRAN}, 
the critic $q^i_\phi$ computes the $Q$-value for an agent $i$
by taking all agents' GRU outputs, 
all other agents' one-hot encoded actions,
and the one-hot encoded agent ID $i$ as input.
The critic contains four successive linear layers with 64 neurons each
and ReLU activations between them. 
The greedy difference $v_\psi$ 
also conditions on all agents' GRU outputs 
and uses three successive linear layers with 64 neurons each
and ReLU activations between them.
After some coarse hyper-parameter exploration for \texttt{QTRAN}
with $\lambda_\text{\sc opt}, \lambda_\text{\sc nopt} \in \{0.1, 1, 10\}$,
we chose the loss parameters 
$\lambda_\text{\sc opt} = 1, \lambda_\text{\sc nopt} = 10$.
The \texttt{LRQ} results in the main text used 
the state-encoding from \texttt{CG} and $K=64$.

All algorithms were trained with one RMSprop gradient step
after each observed episode based on a batch
of 32 episodes, which always contains the newest, 
from a replay buffer holding the last 500 episodes.
The optimizer uses learning rate 0.0005, $\alpha=0.99$ 
and $\epsilon=0.00001$.
Gradients with a norm $\geq 10$ were clipped.
The target network parameters were replaced 
by a copy of the current parameters every 200 episodes.


\begin{figure}[t]
	\includegraphics[width=.48\columnwidth]{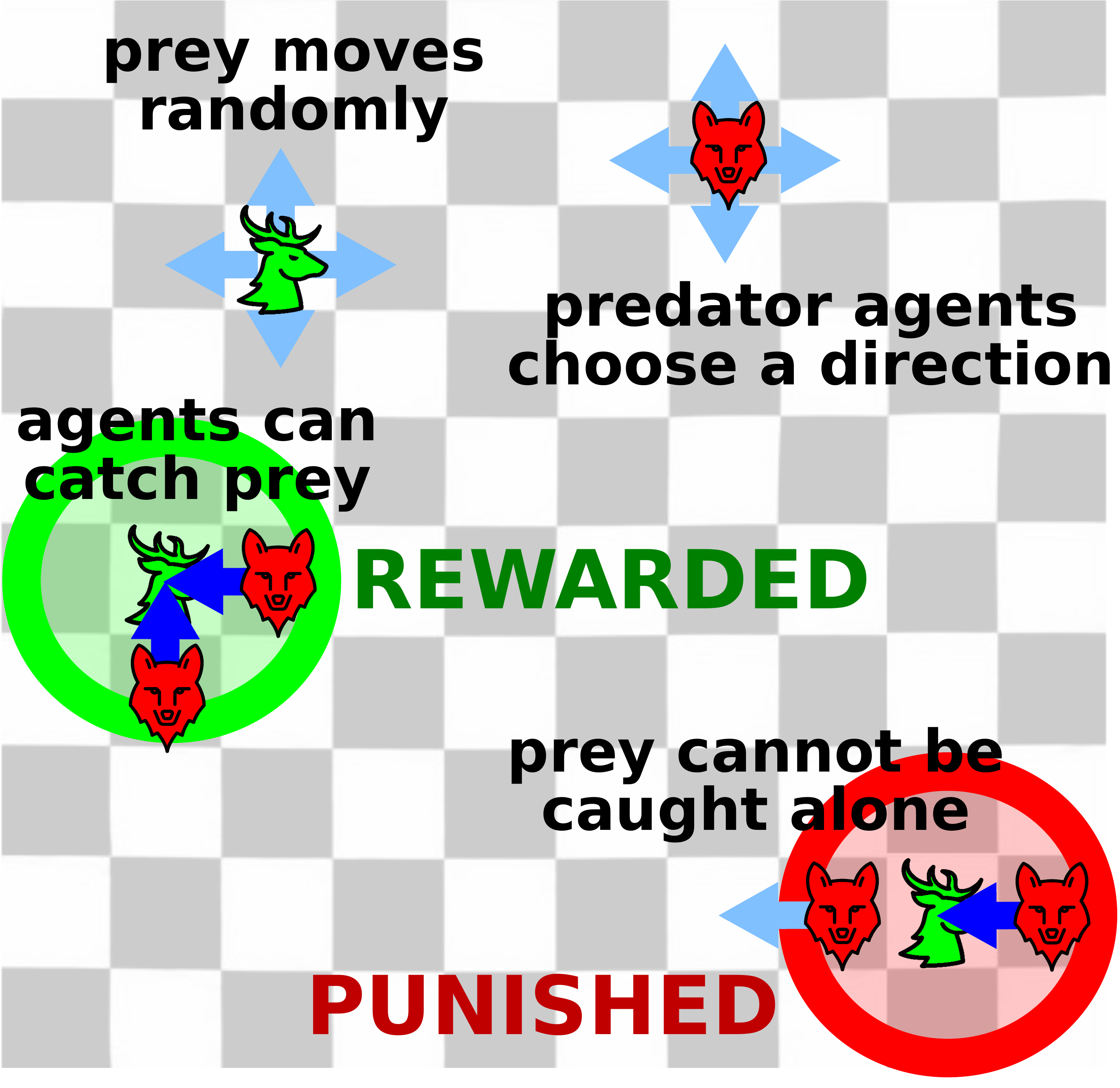}
	\hfill
	\includegraphics[width=.48\columnwidth]{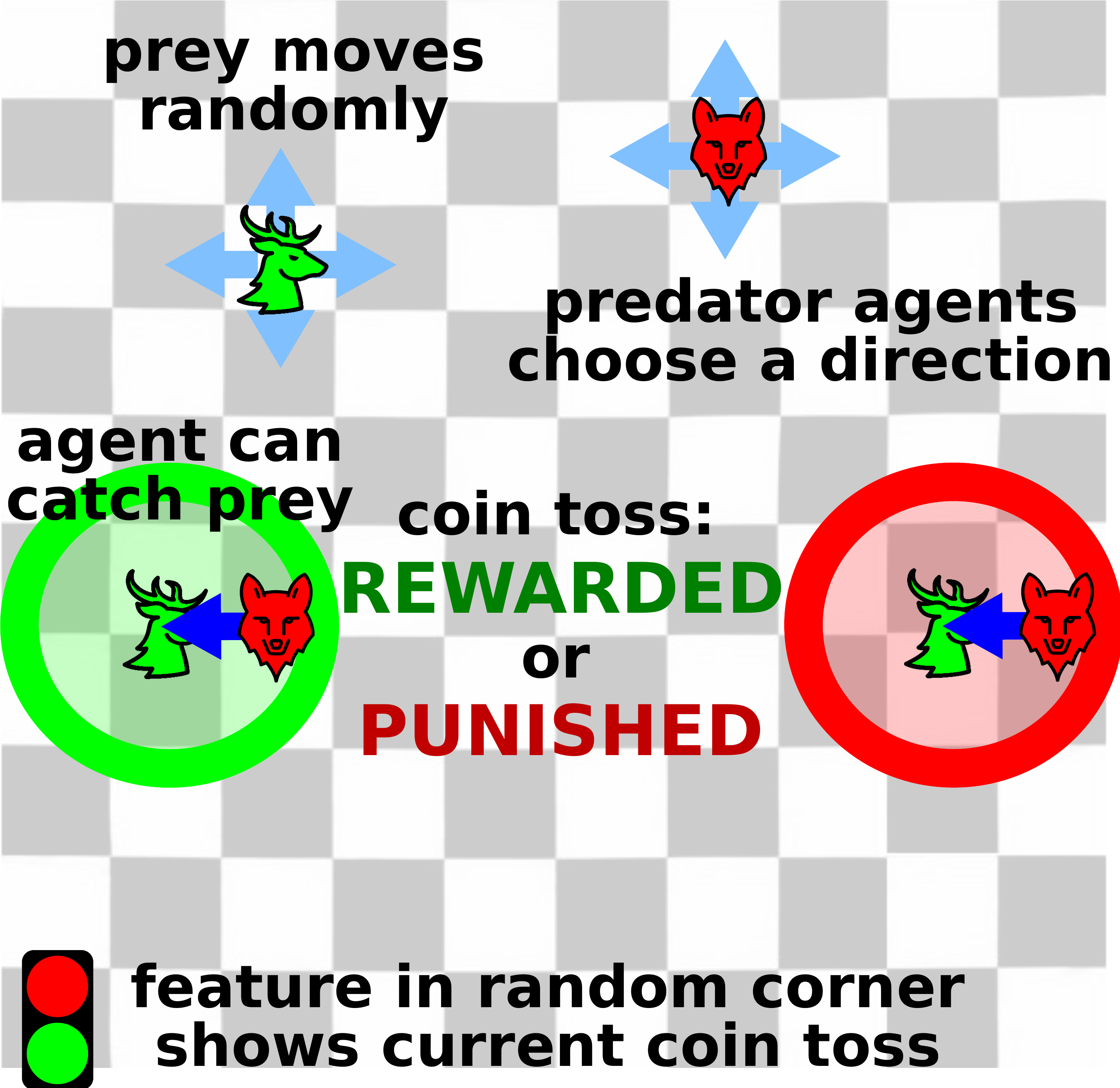}
	\vspace{-2mm}
	\caption{Illustrations of the {\em relative overgeneralization task} 
			(left, Sec.~\ref{sec:rel_overgen})
			and the {\em artificial decentralization task} 
			(right, Sec.~\ref{sec:art_decentral}).}
\end{figure}

\begin{figure*}[t!]
	\vspace{4mm}
	\includegraphics[width=\textwidth]{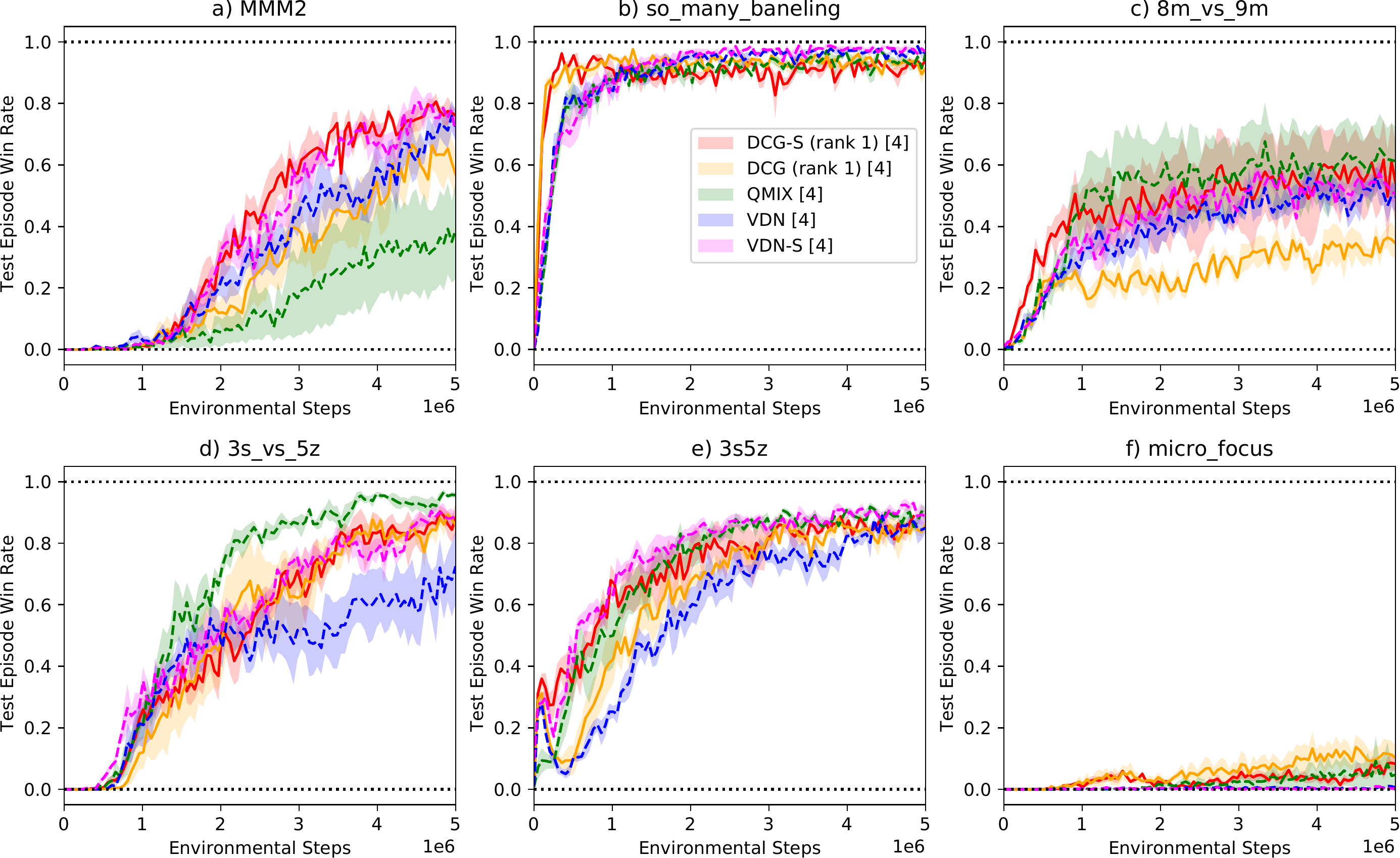}
	\vspace{-4mm}
	\caption{\label{fig:all_starcraft}
			Cumulative reward for test episodes on SMAC maps
			(mean and shaded standard error, [number of seeds])
			for \texttt{QMIX}, \texttt{VDN}, \texttt{VDN-S} 
			and fully connected DCG with 
			rank $K=1$ payoff approximation (\texttt{DCG (rank 1)})
			and additional state-dependent bias function 
			(\texttt{DCG-S (rank 1)}).}
\end{figure*}

\begin{table}[b!]
\def\tabbreak{\\}
\def\arraystretch{1.5}
\def\superhard{{\def\arraystretch{1.0}\!\begin{tabular}{c}super\\hard\end{tabular}\!}}
\begin{center}
\setlength{\tabcolsep}{2pt}
\begin{tabular}{|c|llc|} 
	\hline
	Name & Agents & Enemies & Diff. \tabbreak
	\hline
	\texttt{so\_many\_baneling} & 7 Zealots & 32 Banelings & easy \tabbreak
	\texttt{8m\_vs\_9m} & 8 Marines & 9 Marines & hard \tabbreak
	\texttt{3s\_vs\_5z} & 3 Stalker & 5 Zealots & hard \\[1mm]
	\texttt{3s5z}		& \!\!\!{\def\arraystretch{1.0}\begin{tabular}{l} 
							\,3 Stalker \\ \,5 Zealots \end{tabular}}
						& {\def\arraystretch{1.0}\begin{tabular}{l} 
							\!3 Stalker \\ \!5 Zealots \end{tabular}}
						& hard \\[3mm]
	\texttt{MMM2} 		& \!\!{\def\arraystretch{1.0}\begin{tabular}{l} 
							1 Medivac \\ 2 Marauders \\ 7 Marines \end{tabular}}
		& {\!\!\!\def\arraystretch{1.0}\begin{tabular}{l} 
		1 Medivac \\ 3 Marauders \\ 8 Marines \end{tabular}} & \superhard \tabbreak
	\texttt{micro\_focus}&6 Hydralisks & 8 Stalker & \superhard \tabbreak
	\hline
\end{tabular}
\end{center}
\vspace{-2mm}
\caption{\label{tab:sc2_maps}Types of agents, enemies and difficulty 
	of all tested StarCraft II maps for SMAC \citep{Samvelyan19}.}
\end{table}

\subsection{StarCraft II details} \label{sec:starcraft}

We kept all hyper-parameters the same and evaluated the 
six maps in Table \ref{tab:sc2_maps}.
All maps are from SMAC \citep{Samvelyan19},
except \texttt{micro\_focus},
which was provided to us by the SMAC authors.
The results for \texttt{DCG-S}, \texttt{DCG}, 
\texttt{QMIX} and \texttt{VDN} are given 
in Figure \ref{fig:all_starcraft} on Page \pageref{fig:all_starcraft},
where both DCG variants use a rank-1 payoff approximation.
Note that our results differ from those in \citet{Samvelyan19},
due to slightly different parameters and 
an update after every episode.
The latter differs from the original publication 
because we use the the \texttt{episode\_runner} instead of 
the \texttt{parallel\_runner} of {\sc pymarl}.
These choices ended up improving the performance of QMIX significantly.

As expected, a direct comparison 
with the state-of-the-art method \texttt{QMIX}
depends strongly on the StarCraft II map.
On the one hand, \texttt{DCG-S} clearly outperforms \texttt{QMIX} 
on \texttt{MMM2} (Figure \ref{fig:all_starcraft}a), 
which is classified as {\em super hard} by SMAC.
We also learn much faster on the {\em easy} map 
\texttt{so\_many\_baneling} (Figure \ref{fig:all_starcraft}b).
On the other hand,
\texttt{QMIX} performs better on the {\em hard} map 
\texttt{3s\_vs\_5z} (Figure \ref{fig:all_starcraft}d), 
which might be due to the low number of 3 agents.
For that amount of agents, 
the added representational capacity of DCG may not improve the task
as much as the non-linear state-dependent mixing of QMIX.
It is hard to pin-point why state dependent mixing 
is an advantage here, though.
However, given that \texttt{DCG-S} and \texttt{VDN-S}
perform equally well on all maps except \texttt{so\_many\_baneling}
indicates that the SMAC benchmark probably does not
suffer much from the relative overgeneralization pathology.

\input{core/alg_annotate}
\input{core/alg_qvalue}

\input{core/alg_greedy}

%% file: core/alg_annotate.tex
\begin{algorithm*}[p]
	\begin{algorithmic}
	\Function{annotate}{$\{\ve h_{t-1}^i, a_{t-1}^i, o_t^i \}_{i=1}^n, 
				\Set E, \{\Set A^i\}_{i=1}^{n}, K \in \N$}
			\Comment{$A := |\cup_i \Set A^i|$}
		\State $\ve f^\text{V} := \ve 0 \quad\in\quad \R^{n \times A}$
			\Comment{initialize utility tensor}
		\State $\ve f^\text{E} \hspace{.25mm}:= 
				\ve 0\quad\in\quad \R^{|\Set E| \times A \times A}$
			\Comment{initialize payoff tensor}
		\For {$i \in \{1,\ldots,n\}$}
				\Comment{compute batch with all agents}
			\State $\ve h_t^i := h_\psi(\ve h_{t-1}^i, o_t^i, a_{t-1}^i)$
				\Comment{new hidden state}
			\State $\ve f^\text{V}_{i} \leftarrow f^v_\theta(\ve h^i_t) 
				\quad\in\quad \R^A$
				\Comment{compute utility}
			\For {$a \in \{1,\ldots,A\} \setminus \Set A^i$}
					\Comment{set unavailable actions ...}
				\State $\ve f^\text{V}_{ia} \leftarrow -\infty$
					\Comment{...  to $-\infty$}
			\EndFor
		\EndFor
		\For {$e = (i,j) \in \Set E$}
				\Comment{compute batch with all edges} 
			\If {$K = 0$} \Comment{if no low-rank approximation}
				\State $\ve f^\text{E}_e \leftarrow \smallfrac{1}{2}
					f^e_{\phi}(\cdot, \cdot|\ve h_t^i, \ve h_t^j) 
					+ \smallfrac{1}{2}
					f^e_{\phi}(\cdot, \cdot|\ve h_t^j, \ve h_t^i)^\top
					\;\; \in \;\; \R^{A \times A}$
					\Comment{symmetric payoffs}
			\Else \Comment{if low-rank approximation}
				\State $[\hat{\mat F}, \bar{\mat F}] := 
					f^e_{\phi}(\cdot, \cdot, \cdot|\ve h_t^i, \ve h_t^j)
					\quad\;\; \in \quad \R^{2 \times A \times K}$
				\State $[\hat{\mat F}', \bar{\mat F}'] := 
					f^e_{\phi}(\cdot, \cdot, \cdot|\ve h_t^j, \ve h_t^i)
					\quad \in \quad \R^{2 \times A \times K}$
				\State $\ve f^\text{E}_e \leftarrow 
					\smallfrac{1}{2} \hat{\mat F} \bar{\mat F}^\top
					+ \smallfrac{1}{2} \bar{\mat F}' \hat{\mat F}'^\top
					\qquad \in \quad \R^{A \times A}$
					\Comment{symmetric payoffs}
			\EndIf
		\EndFor
		\State\Return \; $\{\ve h^i_t\}_{i=1}^n, \ve f^\text{V}, \ve f^\text{E}$
			\Comment{return hidden states $\ve h^i_t$, 
					utility tensor $\ve f^\text{V}$ and payoff tensor $\ve f^\text{E}$}
	\EndFunction 
	\end{algorithmic}
	\caption{\label{alg:annotate}
			Annotates a CG by computing the utility 
			and payoff tensors (rank $K$ approximation).} 
\end{algorithm*}

%% file: core/alg_qvalue.tex
\begin{algorithm*}[p]
	\begin{algorithmic}
	\Function{qvalue}{$\ve f^\text{V} \in \R^{|\Set V| \times A}, 
					\ve f^\text{E} \in \R^{|\Set E| \times A \times A},
					\ve a \in \Set A, s_t \in \Set S \cup \{\varnothing\}$} 
		\Comment $v_\varphi(\varnothing) = 0$
		\State	\Return \;$\smallfrac{1}{|\Set V|} \smallsum{i=1}{|\Set V|} 
					\ve f^\text{V}_{ia^i}
				+ \smallfrac{1}{|\Set E|} 
					\hspace{-3mm}\smallsum{e=(i,j) \in \Set E}{}\hspace{-3mm}
					\ve f^\text{E}_{ea^i\!a^j}
				+ v_\varphi(s_t)$ 
			\Comment{return the Q-value of the given actions $\ve a$}
	\EndFunction
	\end{algorithmic}
	\caption{\label{alg:qvalue}
			Q-value computed from utility and payoff tensors 
			(and potentially global state $s_t$).} 
\end{algorithm*}

%% file: core/alg_greedy.tex
\begin{algorithm*}[p]
	\begin{algorithmic}
	\Function{greedy}{$\ve f^\text{V} \in \R^{|\Set V| \times A}, 
					\ve f^\text{E} \in \R^{|\Set E| \times A \times A},
					\Set V, \Set E, \{\Set A^i\}_{i=1}^{|\Set V|}, k
					$}
			\Comment{$A := |\cup_i \Set A^i|$}
		\State $\ve\mu^0, \bar{\ve\mu}^0 := \ve 0 \;\in\; \R^{|\Set E| \times A}$
			\Comment{messages forward ($\ve \mu$) 
					 and backward ($\bar{\ve \mu}$)}
		\vspace{0.5mm}
		\State $\ve q^0 := \frac{1}{|\Set V|} \ve f^\text{V}$
			\Comment{initialize ``Q-value'' without messages}
		\vspace{-0.5mm}
		\State $q_\text{max} := -\infty; \quad 
				  \ve a_\text{max} := \big[\argmax\limits_{a \in \Set A^i} q^0_{ai} 
				  							  \,\big	|\, i \in \Set V \big]$
			\Comment{initialize best found solution}
		\vspace{-1.25mm}
		\For{ $t \;\in\; \{1, \ldots, k\}$ }  
				\Comment{loop with $k$ message passes}
			\vspace{0.5mm}
			\For{ $e = (i,j) \;\in\; \Set E$ }
				\Comment{update forward and backward messages}
				\State $\ve \mu_e^{t} := \max\limits_{a \in \Set A^i} 
						\big\{ (q_{ia}^{t-1} \!- \bar{\mu}_{ea}^{t-1}) 
						+ \frac{1}{|\Set E|} \ve f^\text{E}_{ea} \big\}$
					\Comment{forward: maximize sender}
				\State $\bar{\ve \mu}_e^{t} := \max\limits_{a \in \Set A^j} 
						\big\{ (q_{ja}^{t-1} \!- {\mu}_{ea}^{t-1}) 
						+ \frac{1}{|\Set E|} (\ve f^\text{E}_{e})^{\!\top}_{a} \big\}$
					\Comment{backward: maximizes receiver}
				\If{ \texttt{message\_normalization} }
					\Comment{to ensure converging messages}
					\vspace{1mm}
					\State $\ve \mu^{t}_e \leftarrow \ve \mu^{t}_e 
							- \frac{1}{|\Set A^j|} \sum\limits_{a \in \Set A^j} 
								\mu^t_{ea}$
						\Comment{normalize forward message}
					\State $\bar{\ve \mu}^{t}_e \leftarrow \bar{\ve \mu}^{t}_e 
							- \frac{1}{|\Set A^i|} \sum\limits_{a \in \Set A^i} 
								\bar\mu^t_{ea}$
						\Comment{normalize backward message}
				\EndIf
			\EndFor
			\For{$i \;\in\; \Set V$}
				\Comment{update ``Q-value'' with messages}
				\State $\ve q_i^t := \frac{1}{|\Set V|} \ve f^\text{V}_i 
						\;+\; \hspace{-4mm} 
							{\displaystyle\sum_{e = (\cdot,i) \in \Set E}} 
							\hspace{-3mm} \ve\mu^t_e
						\;+\; \hspace{-4mm}
							{\displaystyle\sum_{e = (i,\cdot) \in \Set E}} 
							\hspace{-3mm} \bar{\ve\mu}^t_e$
					\Comment{utility plus incoming messages}
				\State $a^t_i := \argmax\limits_{a \in \Set A^i} \{ q^t_{ia} \}$
					\Comment{select greedy action of agent $i$}
			\EndFor
			\State $q' \leftarrow \text{\sc qvalue}(\ve f^\text{V}, 
					\ve f^\text{E}, \ve a^t, \varnothing)$
					\Comment{get true Q-value of greedy actions}
			\vspace{0.5mm}
			\If{$q' > q_\text{max}$}
				$\{ \ve a_\text{max} \leftarrow \ve a^t; \;\; 
						q_\text{max} \leftarrow q'\}$
					\Comment{remember only the best actions}
			\EndIf
		\vspace{1mm}
		\EndFor
		\Return \;$\ve a_\text{max} \in \Set A^1 \times \ldots \times \Set A^{|\Set V|}$ 
			\Comment{return actions that maximize the joint Q-value}
	\EndFunction
	\end{algorithmic}
	\caption{\label{alg:greedy}
		Greedy action selection with $k$ message passes in a coordination graph.} 
\end{algorithm*}